\documentclass{article}

 \usepackage[preprint]{neurips_2025}

\usepackage[utf8]{inputenc} 
\usepackage[T1]{fontenc}    
\usepackage{url}            
\usepackage{booktabs}       
\usepackage{amsfonts}       
\usepackage{nicefrac}       
\usepackage{microtype}      
\usepackage{graphicx}
\usepackage{wrapfig}
\usepackage{subcaption}
\usepackage[table]{xcolor}        
\usepackage{tabularx}      
\usepackage{adjustbox}     
\usepackage{booktabs}      
\usepackage{array}         
\usepackage{multirow}      
\usepackage{makecell}      
\usepackage{tablefootnote}
\usepackage[symbol]{footmisc}
\usepackage{soul}
\usepackage{amsmath} 
\usepackage{amsfonts} 
\usepackage{amssymb} 
\usepackage{tcolorbox}
\usepackage{listings} 
\tcbuselibrary{listings}

\newtcblisting{promptbox}[2][]{
  title={\textbf{#2}},
  colback=gray!5,
  colframe=gray!50,
  fonttitle=\bfseries,
  listing only,
  listing options={breaklines=true, basicstyle=\ttfamily\small},
  #1
}

\usepackage[pagebackref=true, breaklinks=true, colorlinks,
            citecolor=citecolor, linkcolor=linkcolor, bookmarks=false]{hyperref}
\definecolor{citecolor}{HTML}{0071BC}
\definecolor{linkcolor}{HTML}{ED1C24}

\usepackage{amssymb}
\usepackage{pifont}

\usepackage{multibib}
\newcites{app}{Appendix References}

\definecolor{c1}{RGB}{24, 209, 207}
\definecolor{c2}{RGB}{22, 195, 203}
\definecolor{c3}{RGB}{20, 182, 198}
\definecolor{c4}{RGB}{18, 168, 194}
\definecolor{c5}{RGB}{16, 154, 189}
\definecolor{c6}{RGB}{14, 140, 185}
\definecolor{c7}{RGB}{12, 126, 180}
\definecolor{c8}{RGB}{10, 112, 176}
\definecolor{c9}{RGB}{8, 52, 154}

\definecolor{spatial}{RGB}{230,242,255} 
\definecolor{planning}{RGB}{255,240,230} 
\definecolor{partial}{RGB}{230,255,230} 
\definecolor{temporal}{RGB}{245,230,255} 
\definecolor{header}{RGB}{220,230,242} 
\definecolor{lightrow}{RGB}{245,247,250} 

\title{\textcolor{c1}{T}\textcolor{c2}{e}\textcolor{c3}{x}\textcolor{c4}{t}\textcolor{c5}{A}\textcolor{c6}{t}\textcolor{c7}{a}\textcolor{c8}{r}\textcolor{c9}{i}: 100K Frames Game Playing with Language Agents}

\author{%
  Wenhao Li$^{1}$ \quad
  Wenwu Li$^{1}$ \quad
  Chuyun Shen$^{2}$ \quad
  Junjie Sheng$^{3}$ \quad
  Zixiao Huang$^{2}$ \quad
  Di Wu$^{1}$ \AND
  \vspace{-10pt}
  Yun Hua$^{4}$ \quad
  Wei Yin$^{5}$ \quad
  Xiangfeng Wang$^{2}$ \quad
  Hongyuan Zha$^{6}$ \quad
  Bo Jin$^{1}$ \quad
  \\
  \\
  \\
  $^1$ Tongji University, Shanghai, China \ 
  $^2$ East China Normal University, Shanghai, China \\
  $^3$ Independent Researcher, Shanghai, China \\
  $^4$ Shanghai Jiao Tong University, Shanghai, China \ 
  $^5$ Bank of Communications, Shanghai, China\\
  $^6$ The Chinese University of Hong Kong, Shenzhen, China\\
  \texttt{\{whli, wenwu, wu2002, bjin\}@tongji.edu.cn}, \ 
  \texttt{zhahy@cuhk.edu.cn} \\
  \texttt{\{jarvis@stu, zxhuang@stu, cyshen@stu, xfwang@cs\}.ecnu.edu.cn} \\
  \texttt{yinw\_8@bankcomm.com}, \
  \texttt{hyyh28@sjtu.edu.cn}
}

\begin{document}

\maketitle

\begin{figure}[htb!]
    \centering
    \includegraphics[width=\linewidth]{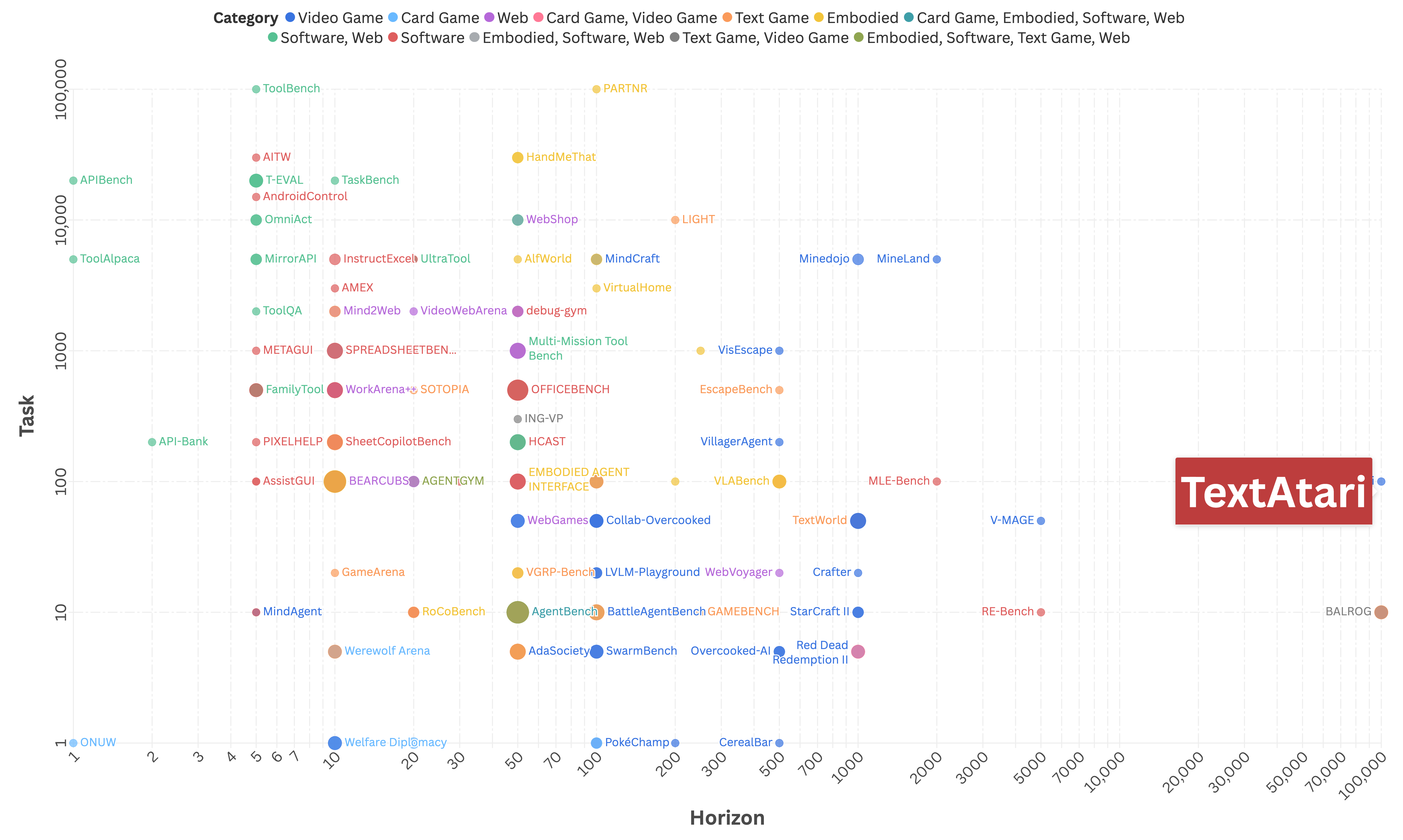}
    \caption{Statistics of tasks and horizons.}
    \label{fig:enter-label}
\end{figure}

\begin{abstract}
  We present TextAtari, a comprehensive benchmark for evaluating language agents on very long-horizon decision-making tasks spanning up to $100,000$ steps. 
  By translating the visual state representations of classic Atari games into rich textual descriptions, TextAtari creates a challenging test bed that bridges sequential decision-making with natural language processing. 
  Our benchmark encompasses nearly 100 distinct tasks with varying complexity, action spaces, and planning horizons, all rendered as text through an unsupervised representation learning framework (AtariARI). 
  We evaluate three open-source large language models (Qwen2.5-7B, Gemma-7B, and Llama3.1-8B) across three agent frameworks (zero-shot, few-shot chain-of-thought, and reflection reasoning) to systematically assess how different forms of prior knowledge affect performance on these unprecedented long-horizon challenges. 
  The four distinct scenarios—Basic, Obscured, Manual Augmentation, and Reference-based—investigate the impact of semantic understanding, instruction comprehension, and expert demonstrations on agent decision-making. 
  Our results reveal significant performance gaps between language agents and human players in these extensive planning tasks, highlighting challenges in sequential reasoning, state tracking, and strategic planning across tens of thousands of steps. TextAtari provides standardized evaluation protocols, baseline implementations, and a comprehensive framework for advancing research at the intersection of language models and planning.
  Our code is available at~\url{https://github.com/Lww007/Text-Atari-Agents}.
\end{abstract}

\section{Introduction}\label{sec:intro}

\begin{wrapfigure}{r}{0.5\linewidth}
    \centering
    \includegraphics[width=\linewidth]{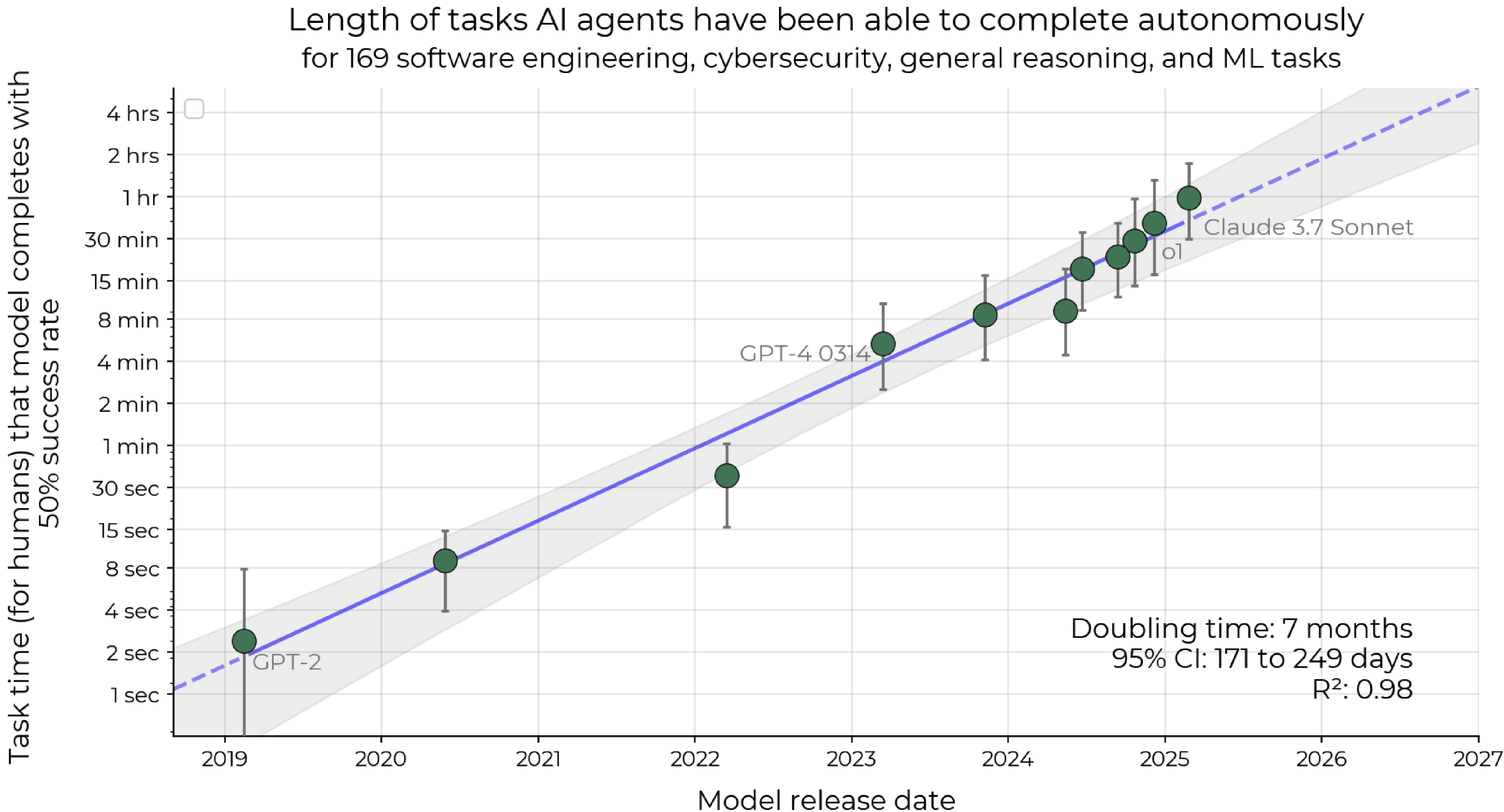}
    \caption{Screenshot from~\citet{kwa2025measuring}.}
    \label{fig:trend}
\end{wrapfigure}

Sequential decision-making over extended time horizons represents one of the most fundamental challenges in artificial intelligence~\citep{aghzal2025survey,kang2024empirical,pignatelli2023survey}. 
While humans naturally navigate complex, long-term planning scenarios—from playing strategic games to coordinating daily activities—AI systems have traditionally struggled with tasks requiring thousands of interdependent decisions. 
Recent analyses reveal a striking trend: 
the length of tasks that generalist autonomous AI agents can complete with $50$\% reliability has been doubling approximately every $7$ months for the past $6$ years~\citep{kwa2025measuring}, as shown in Figure~\ref{fig:trend}. 
This exponential growth suggests that within a decade, AI agents may independently complete tasks that currently take humans days or weeks—yet a critical gap remains in our ability to evaluate these systems on truly long-horizon challenges.

\begin{wrapfigure}{l}{0.4\linewidth}
    \centering
    \includegraphics[width=\linewidth]{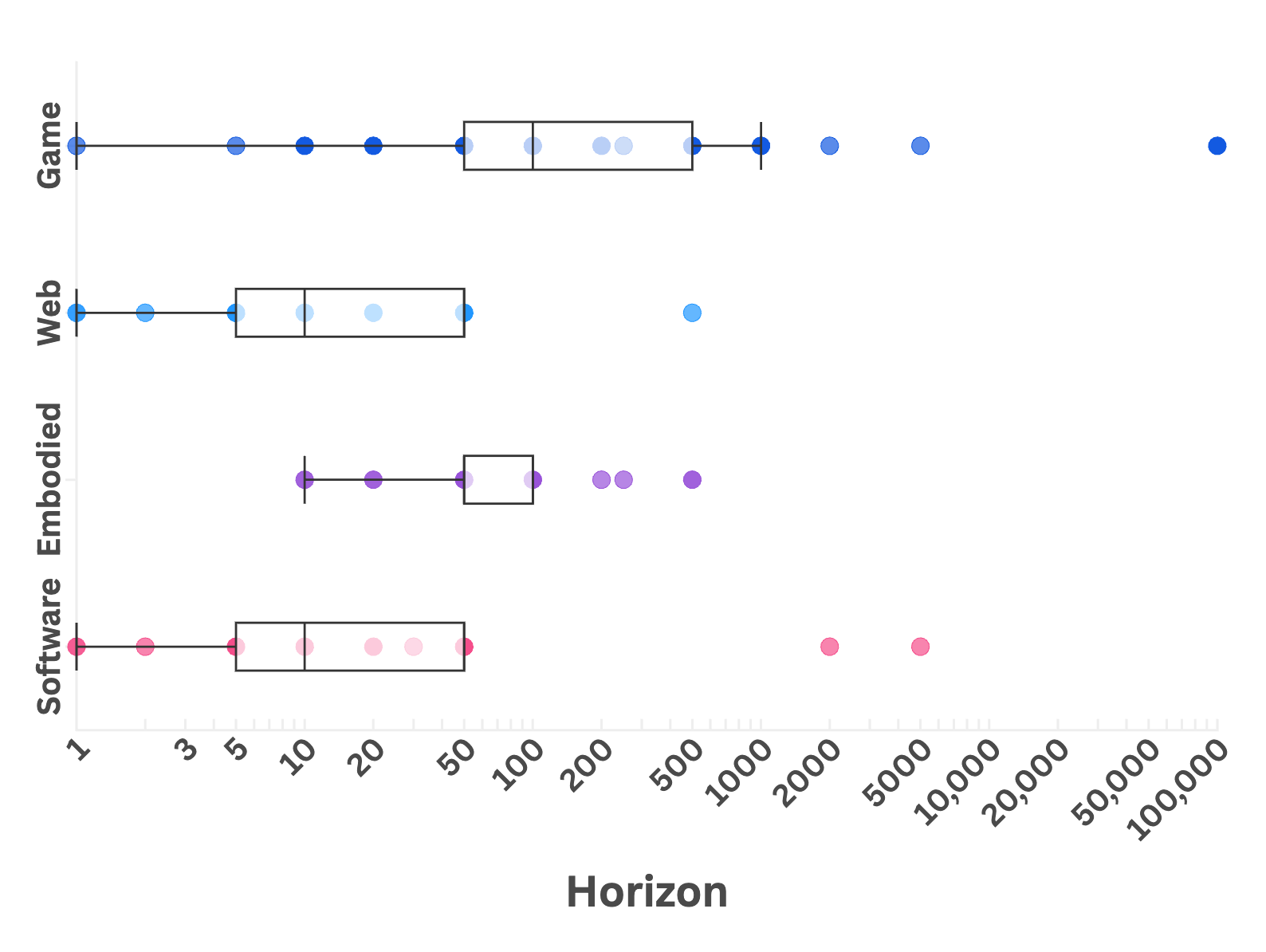}
    \caption{Horizon statistics.}
    \label{fig:horizon-stat}
\end{wrapfigure}

The AI community has developed numerous benchmarks for evaluating sequential decision-making, spanning web interfaces, desktop software, games, and embodied environments~\citep{tan2024cradle}. 
However, our comprehensive analysis of $163$ existing sequential decision benchmarks (see Appendix for the full list) reveals a critical limitation: most operate on remarkably short horizons. 
While web and software manipulation tasks typically involve fewer than $50$ steps, text and video games represent the longest horizon challenges ($75$\% approach $500$ steps)—yet even these remain dramatically underexplored, with only $4$ out of $163$ benchmarks reaching the $100,000$-step threshold\footnote{According to estimates from previous work~\citep{kwa2025measuring}, current large language model agents can complete tasks that take humans less than $4$ minutes with a success rate approaching $100$\%. Based on the statistics compiled in this paper (as shown in Figure~\ref{fig:horizon-stat}), the median number of decision steps in most benchmarks is around $10$ steps. Therefore, $100,000$ steps would correspond to tasks that would take humans about a week to complete, this would require approximately $2$-$4$ years of development time~\citep{kwa2025measuring}, making the $100,000$-step challenge an appropriately difficult goal that is unlikely to be achieved in the near future.} needed to evaluate truly extended reasoning\footnote{These benchmarks with ultra-long decision-making sequences have their own limitations, which will be discussed shortly.}.

This horizon gap is particularly concerning as digital games emerge as the most challenging sequential decision domains due to their unique combination of environmental complexity, non-linear decision paths, and partial observability—requiring agents to store and reason upon past experiences for effective decision-making~\citep{ecoffet2019go,fan2022minedojo,ma2024large}. 
As language models increasingly serve as the cognitive engine for autonomous systems, their capacity for sustained reasoning and decision-making across very long horizons becomes a critical frontier for research.

Recent advances in large language models (LLMs)~\citep{anthropic2025claude37sonnet,guo2025deepseek,jaech2024openai} have demonstrated remarkable capabilities in reasoning, planning, and decision-making within short contexts. 
These models can generate coherent multi-step plans and engage in complex reasoning tasks~\citep{chen2025towards,li2025system}.
However, their ability to maintain consistent decision-making over very long horizons—spanning tens of thousands of steps—remains largely unexplored. 
This represents not merely a quantitative challenge but a qualitative shift in the nature of reasoning required: 
from short-term tactical decisions to long-term strategic planning with compounding consequences.

Consider the challenge of playing a video game. 
Human players naturally track game state, form strategic plans, adapt to changing conditions, and execute thousands of actions over extended gameplay sessions. 
This requires maintaining contextual awareness, making predictions about future states, and continuously adjusting strategies based on feedback—all capabilities essential for general-purpose AI systems~\citep{lake2015human,lake2017building,rips2015divisions}. 
Yet current language agents struggle with such long-horizon tasks, particularly when visual information must be processed through language descriptions rather than direct observation.

In this work, we introduce TextAtari, a comprehensive benchmark for evaluating language agents on very long-horizon decision-making tasks spanning up to $100,000$ steps. 
TextAtari transforms the visual states of classic Atari games into rich textual descriptions using an unsupervised representation learning framework (AtariARI)~\citep{anand2019unsupervised}, creating a challenging testbed that bridges sequential decision-making with natural language processing. 
Our decision to transform visual game environments into textual representations is methodologically motivated. 
While visual-language models have advanced significantly, they remain limited in reasoning depth, context length handling, and decision coherence compared to text-only language models~\citep{yang2022dichotomy,yang2024video}. 
This approach enables evaluation of pure reasoning capabilities by eliminating confounding variables associated with visual processing, allows precise control over information representation, provides standardized inputs for enhanced experimental reproducibility, and aligns with the theoretical framework of language models as general reasoning engines interacting with environments through symbolic descriptions. 
TextAtari thus establishes a controlled experimental environment for evaluating language agents' capacity to maintain coherent reasoning across extended horizons.

Our benchmark encompasses nearly $100$ distinct tasks with varying complexity, action spaces, and planning horizons. 
TextAtari offers several key advantages: 
First, Atari games provide well-defined environments with clear objectives and measurable performance metrics. 
Second, by rendering these traditionally visual environments as text, we can directly evaluate language agents' ability to process, reason about, and act upon textual information over extremely long horizons. 
Third, our four distinct scenario designs—Basic, Obscured, Manual Augmentation, and Reference-based—enable systematic investigation of how different forms of prior knowledge affect agent performance.

We hope TextAtari will serve as a valuable resource for the research community, providing standardized evaluation protocols, baseline implementations, and a comprehensive framework for advancing research at the intersection of language models and planning. 
Progress on this benchmark could lead to language agents capable of coherent decision-making across very long time horizons, bringing us closer to AI systems that can maintain consistent reasoning and planning at human-like scales.

\section{Benchmark Design and Construction}\label{sec:design}

TextAtari addresses the critical gap in existing sequential decision-making evaluations by transforming visual Atari environments into rich textual descriptions for language model processing. 
Our benchmark targets horizons of up to $100,000$ steps—a threshold reached by fewer than $2.5$\% of existing benchmarks. 
We employed AtariARI, an unsupervised representation learning framework, to convert visual states into detailed textual descriptions that preserve essential gameplay features while creating a challenging testbed for language-based reasoning.
The benchmark construction process involved selecting Atari games that represent diverse challenges in sequential decision-making. 
We prioritized games with varying complexity levels, action spaces, and planning horizons to evaluate different aspects of language agents' long-horizon reasoning capabilities.

\subsection{Task Suite}

Our task suite encompasses $23$ classic Atari games spanning four major categories: 
Action Games, Puzzle and Strategy Games, Sports Games, and Arcade Classics. 
Each category presents distinct challenges for language models, testing different aspects of reasoning, planning, and decision-making over extended time horizons.

\begin{table}[!t]
\begin{adjustbox}{center, width=\textwidth}
\small
\begin{tabularx}{\textwidth}{>{\hsize=0.27\hsize}X>{\hsize=0.73\hsize}X}
\rowcolor{header} \multicolumn{2}{c}{\textbf{Detailed Description of Atari Games}} \\
\midrule
\textbf{Game} & \textbf{Detailed Description} \\
\midrule
\rowcolor{lightrow} Venture & An exploration game where players control Winky, an adventurer navigating through a multi-room dungeon to collect treasures. Each room contains different monsters guarding treasure, requiring specific strategies to overcome. Players view the dungeon layout from an overhead perspective but transition to a zoomed-in view when entering a room. If players take too long in a room, the invincible "Hallmonster" appears, forcing swift action. The game features four different dungeons with increasing difficulty and unique monsters in each room, from snakes and trolls to giant spiders and the Grim Reaper. \\
\midrule
VideoPinball & A digital recreation of pinball that simulates the physics and features of a traditional pinball machine. Players control left and right flippers to keep the ball in play, aiming to hit various targets to score points. The table includes bumpers, spinners, rollover targets, and bonus areas. Players can tilt the table (with limits) to influence ball direction. The game features realistic ball physics including momentum, ricochet angles, and speed changes. Special features include multiball play and bonus rounds that can be activated through specific target combinations. Scoring emphasizes both quick reflexes and strategic target selection. \\
\bottomrule
\end{tabularx}
\end{adjustbox}
\caption{
This table provides detailed descriptions of selected atari games., explaining their gameplay mechanics, objectives, and distinctive features.}
\label{table:atari_games_description4-main}
\vspace{-10pt}
\end{table}

Action Games like Asteroids and Berzerk challenge models with spatial reasoning requirements and strategic target prioritization. 
Puzzle and Strategy Games such as Breakout and MontezumaRevenge emphasize planning and trajectory prediction, with MontezumaRevenge featuring extremely sparse rewards that require extensive planning. 
Sports Games including Boxing and Tennis present challenges in adversarial reasoning and anticipating opponent behaviors. 
Arcade Classics like MsPacman and Seaquest demand dynamic path planning and resource management in partially observable environments.
Each game presents unique combinations of challenges across four key dimensions: 
spatial reasoning, planning and strategy, partial observability, and temporal reasoning. 
For instance, Seaquest requires sophisticated resource management (oxygen) while MsPacman demands strategic power pellet usage and ghost behavior modeling.

\begin{table}[htb!]
\begin{adjustbox}{center, width=\textwidth}
\small
\begin{tabularx}{\textwidth}{>{\hsize=0.4\hsize}X>{\hsize=0.7\hsize}X>{\hsize=0.9\hsize}X}
\rowcolor{header} \multicolumn{3}{c}{\textbf{Atari Games Classification and LLM Gaming Challenges (Summary)}} \\
\midrule
\textbf{Game} & \textbf{Category} & \textbf{Challenges for LLM} \\
\midrule
\rowcolor{gray!10} \multicolumn{3}{c}{\textbf{Action Games}} \\
\midrule
Asteroids & Space Shooter & 
\begin{tabular}{p{0.9\hsize}}
\cellcolor{spatial}Spatial reasoning for circular movement \\
\cellcolor{temporal}Reaction-based gameplay timing \\
\cellcolor{planning}Strategic target prioritization
\end{tabular} \\
\midrule
Berzerk & Maze Shooter & 
\begin{tabular}{p{0.9\hsize}}
\cellcolor{spatial}Navigating complex maze layouts \\
\cellcolor{partial}Dynamic obstacle avoidance \\
\cellcolor{planning}Multitasking (walls, enemies, bullets)
\end{tabular} \\
\midrule
\rowcolor{gray!10} \multicolumn{3}{c}{\textbf{Puzzle and Strategy Games}} \\
\midrule
Breakout & Brick-breaker & 
\begin{tabular}{p{0.9\hsize}}
\cellcolor{spatial}Geometry understanding \\
\cellcolor{planning}Trajectory prediction \\
\cellcolor{temporal}Timing-sensitive paddle control
\end{tabular} \\
\midrule
MontezumaRevenge & Puzzle Platformer & 
\begin{tabular}{p{0.9\hsize}}
\cellcolor{planning}Extremely sparse rewards \\
\cellcolor{planning}Complex dependency hierarchies \\
\cellcolor{temporal}Precise timing for trap avoidance
\end{tabular} \\
\midrule
\rowcolor{gray!10} \multicolumn{3}{c}{\textbf{Sports Games}} \\
\midrule
Boxing & Sports & 
\begin{tabular}{p{0.9\hsize}}
\cellcolor{partial}Adversarial reasoning \\
\cellcolor{planning}Tactical positioning \\
\cellcolor{temporal}Timing attack and defense moves
\end{tabular} \\
\midrule
Tennis & Sports & 
\begin{tabular}{p{0.9\hsize}}
\cellcolor{spatial}Court positioning strategy \\
\cellcolor{partial}Opponent behavior anticipation \\
\cellcolor{planning}Shot selection planning
\end{tabular} \\
\midrule
\rowcolor{gray!10} \multicolumn{3}{c}{\textbf{Arcade Classics}} \\
\midrule
MsPacman & Maze & 
\begin{tabular}{p{0.9\hsize}}
\cellcolor{planning}Dynamic path planning \\
\cellcolor{partial}Ghost behavior modeling \\
\cellcolor{planning}Strategic power pellet usage
\end{tabular} \\
\midrule
Seaquest & Underwater Shooter & 
\begin{tabular}{p{0.9\hsize}}
\cellcolor{planning}Resource management (oxygen) \\
\cellcolor{planning}Multi-objective balancing \\
\cellcolor{partial}Bidirectional threat assessment
\end{tabular} \\
\bottomrule
\end{tabularx}
\end{adjustbox}
\caption{
\textbf{Selected Atari Games and Their LLM Challenges (Summary).} This table presents key examples from each game category with their primary challenges for LLMs. Color coding indicates challenge types: \colorbox{spatial}{spatial reasoning}, \colorbox{planning}{planning \& strategy}, \colorbox{partial}{partial observability}, and \colorbox{temporal}{temporal reasoning}.
}
\label{table:atari_games_summary}
\end{table}

\subsection{Environment Generation Pipeline}

\begin{figure}[htb!]
    \centering
    \includegraphics[width=\linewidth]{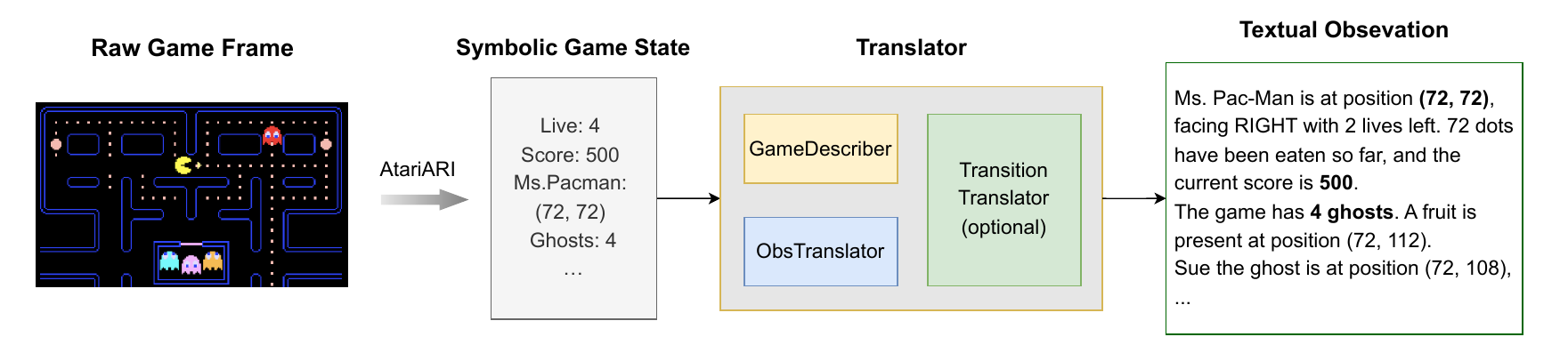}
    \caption{Environment verbalization with AtariARI.}
    \label{fig:pipeline}
\end{figure}

Our framework transforms standard Atari games from the Arcade Learning Environment (ALE) into a purely language-based interface for agents, as shown in Figure~\ref{fig:pipeline}.
Instead of pixel observations, the environment exposes a symbolic game state description at each timestep, allowing a large language model (LLM) to perceive the game through text alone. 
This is accomplished by extracting high-level state variables from the emulator's RAM ($128$ bytes) and converting them into natural language. 
In particular, we leverage the Atari Annotated RAM Interface (AtariARI) wrapper, which provides structured labels for key game entities and variables (e.g. player and object coordinates, scores, lives) by mapping RAM bytes to human-interpretable state information. 
These raw values are then annotated and linearized into textual observations using template-based descriptions. 

Each game is supported by a dedicated translator module following a common design: 
a \texttt{GameDescriber} component provides static context (a brief overview of the game mechanics, objectives, and the action space in words), an \texttt{ObsTranslator} maps each current state vector to a sentence (or set of sentences) describing the agent's situation (for example, reporting positions of relevant objects and the current score), and a \texttt{TransitionTranslator} extends this by narrating state transitions (integrating the last state, the agent's chosen action with a textual label, any reward obtained, and the resulting next state). 
This modular template structure generalizes across games -- new Atari games can be integrated by implementing a similar translator with game-specific vocabulary and templates, while reusing the same interface methods for observations and transitions. 
For instance, the Bowling translator reports the ball and pin positions along with frame scores, whereas the Boxing translator details both fighters' locations and points; 
both adhere to the same class structure and output format. 

To interface smoothly with an LLM-based agent, we throttle the observation generation to a fixed frequency (approximately 5~Hz), ensuring the agent has time to process each description and choose an action. 
We also enforce a limit on the description length (in tokens) so that each observation comfortably fits within the context window of the LLM. 
The result is a pixel-free, symbol-grounded interaction loop: 
the agent perceives an Atari game only through textual narratives of the game state and responds with actions accordingly, enabling direct application of language reasoning and prompting techniques to real-time game control.

To disentangle the contribution of different knowledge sources we define four textual conditions that vary only in the auxiliary information supplied to the language model, while holding the evaluation budget, sampling frequency, and backbone checkpoint fixed. 
Each condition introduces distinct types of prior knowledge into the prompt, allowing us to isolate their individual effects on agent performance. 
Below, we describe each setting in detail and provide the exact prompting format used during interaction.

\noindent\textbf{Basic Scenario.} 
At every step the agent receives only the live textual observation generated by the TextAtari interface. 
Apart from this stream and the immutable header (game synopsis, goals, legal actions), no external material is introduced, making Basic the reference against which all other conditions are measured. 
The prompt includes a static instruction and the latest observation with action choices.

\noindent\textbf{Obscured Scenario.} 
This condition tests reliance on lexical priors by replacing each domain-specific noun in the observation sentence--such as \emph{ghost}, \emph{paddle}, or \emph{asteroid}--with the neutral token \emph{item}. 
The transformation is executed dynamically at runtime using a fixed dictionary (e.g., \texttt{``ghost''} $\rightarrow$ \texttt{``item''}, \texttt{``ball''} $\rightarrow$ \texttt{``item''}). 
This forces the agent to interpret environment dynamics based on structure and positioning rather than familiar words. 
Numbers, coordinates, colours, scores, and the initial header remain untouched.

\noindent\textbf{Manual Augmentation Scenario.} 
To supply explicit rule knowledge, we prepend a concise manual excerpt to the prompt at the start of every episode. 
Manuals are harvested once per game from online repositories such as AtariAge. 
If scanned, the pages are passed through an OCR pipeline, and then summarised by a large language model (e.g., GPT-4) into $\le 300$ tokens capturing key information about controls, scoring mechanics, and game-end conditions. 
This summary is inserted after the game header and serves as grounding for the LLM during gameplay.

\noindent\textbf{Reference-based Scenario.} 
Here the agent is primed with an expert demonstration prior to gameplay. 
For each title, we train a Proximal Policy Optimisation (PPO) controller using Stable-Baselines3 until it reaches at least average human performance. 
A single full evaluation episode is then recorded and subsampled by extracting every 10th state--action pair. 
The state is converted into text via the Text-Atari encoder, and the action is rendered using a task-specific verb template. 
These state--action entries are then concatenated chronologically into a $400$-token trajectory block injected once before live inference begins. 
This method offers the model an in-context exemplar of competent play without directly encoding future knowledge.

\subsection{TextAtari Statistics}

\begin{figure}[htb!]
    \centering
    \includegraphics[width=\linewidth]{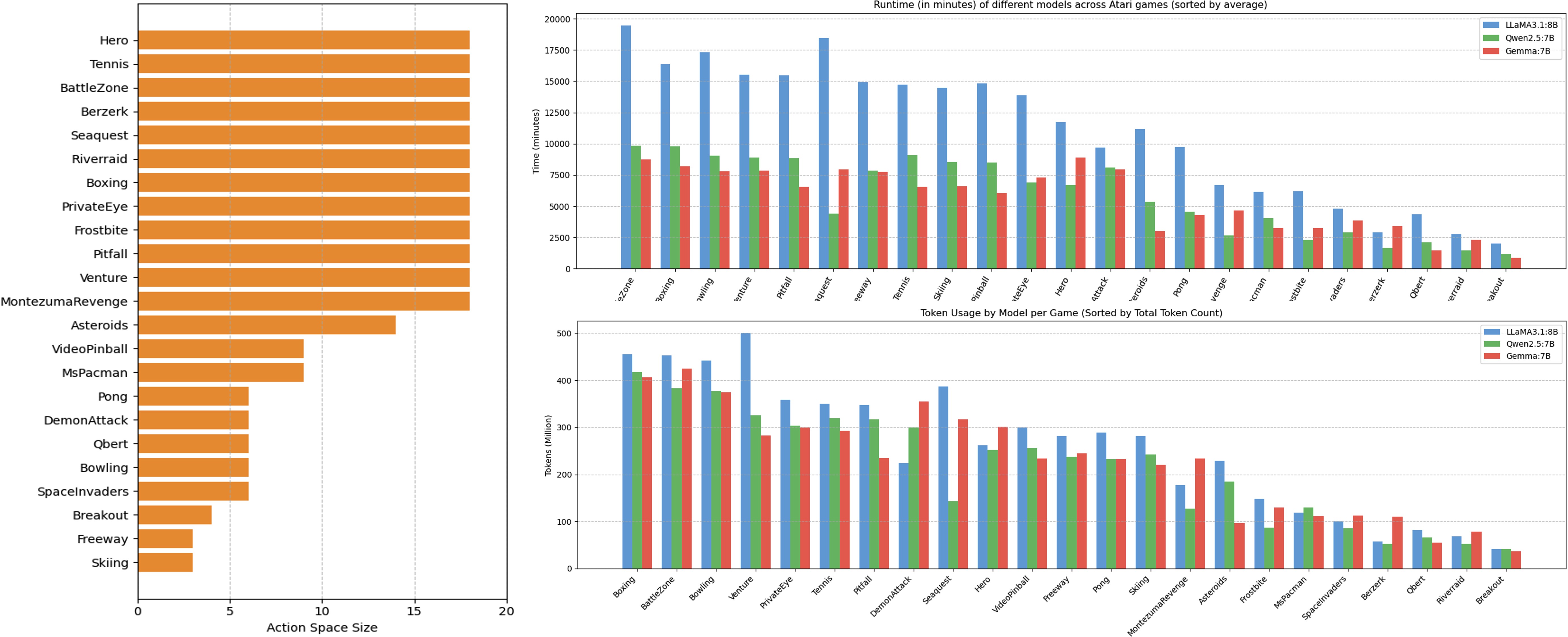}
    \caption{Computational costs.}
    \label{fig:compute}
\end{figure}

Our experiments revealed significant computational demands across the TextAtari benchmark, as shown in Figure~\ref{fig:compute}. 
The left figure illustrates the considerable variation in action space complexity across the $23$ Atari games, with games like Hero, Tennis, and BattleZone exhibiting action spaces approximately three times larger than simpler games such as Skiing and Freeway. 
This action space diversity creates varying degrees of decision complexity that challenge language models differently.

The computational resources required for these experiments were substantial, as shown in the upper right figure. 
Runtime performance varied dramatically across models and games, with LLaMA3.1-8B consistently requiring the most computation time—exceeding $18,000$ minutes ($300$ hours) for multiple games. Qwen2.5-7B and Gemma-7B demonstrated more efficient processing, typically requiring $60$-$70$\% of LLaMA3.1's runtime. The most computationally intensive games (Zaxxon, Seaquest, and Tennis) required over $15,000$ minutes of processing time per model, highlighting the extraordinary computational demands of evaluating long-horizon reasoning.

Token consumption (lower right figure) further emphasizes the scale of these experiments. 
The most token-intensive games consumed between $40$k-$50$k tokens per decision step across all models, with LLaMA3.1-8B consistently using more tokens than its counterparts. 
For context-heavy games like Boxing and BattleZone, this translated to billions of tokens processed throughout the full evaluation. 
This extreme token consumption approaches the practical limits of current context windows, particularly when agents must maintain coherent reasoning across millions steps.

In total, our comprehensive evaluation consumed approximately $820,000$ GPU-minutes on A100 hardware, representing one of the most computationally intensive benchmarks for language agents to date. 
This substantial resource commitment underscores the challenge and importance of evaluating long-horizon reasoning capabilities.

\section{Experiments}\label{sec:exp}

\subsection{Evaluation Protocol}

All scenarios employ the identical inference agent, language model, and roll-out protocol: 
$23$ Atari $2600$ games, a horizon of $1000$ interaction steps (for cost saving), and five random seeds.
Because augmentation increases prompt length, a sliding-window policy discards the oldest system-level messages whenever the projected token count approaches the model's context limit, ensuring that every query remains admissible. 
Consequently, observed performance differences can be attributed to the injected knowledge rather than disparities in prompt size or compute budget.

\subsection{Baselines}

We examine three dialogue policies--Basic, Chain-of-Thought (CoT), and CoT with Reflection--that differ only in the structure and feedback they impose on the language model; 
all external augmentations (manual excerpts, expert trajectories, noun masking) are injected \emph{before} prompt assembly and therefore affect the three agents identically. 
The discussion below focuses exclusively on each agent's internal procedure for constructing, delivering, and updating prompts.

\vspace{0.1cm}
\noindent\textbf{Basic.}
At every decision step the Basic agent issues the leanest prompt possible. It consists only of (i) a static \texttt{system} header that presents a one-sentence synopsis of the game, the win or termination clause, and a numbered list of legal actions, and (ii) a single \texttt{user} message that embeds the live Text-Atari observation. The agent neither requests a reasoning trace nor retains any form of memory; no few-shot examples or auxiliary knowledge are provided. 
The user instruction is fixed across all games and episodes, enforcing a strict zero-shot setting.
The language model must respond with a single valid action identifier--in JSON form when required--without any additional commentary, making this template the shortest baseline against which richer prompting schemes are evaluated.

\begin{wrapfigure}{r}{0.3\textwidth}
    \centering
    \begin{subfigure}[b]{\linewidth}
        \centering
        \includegraphics[width=\linewidth]{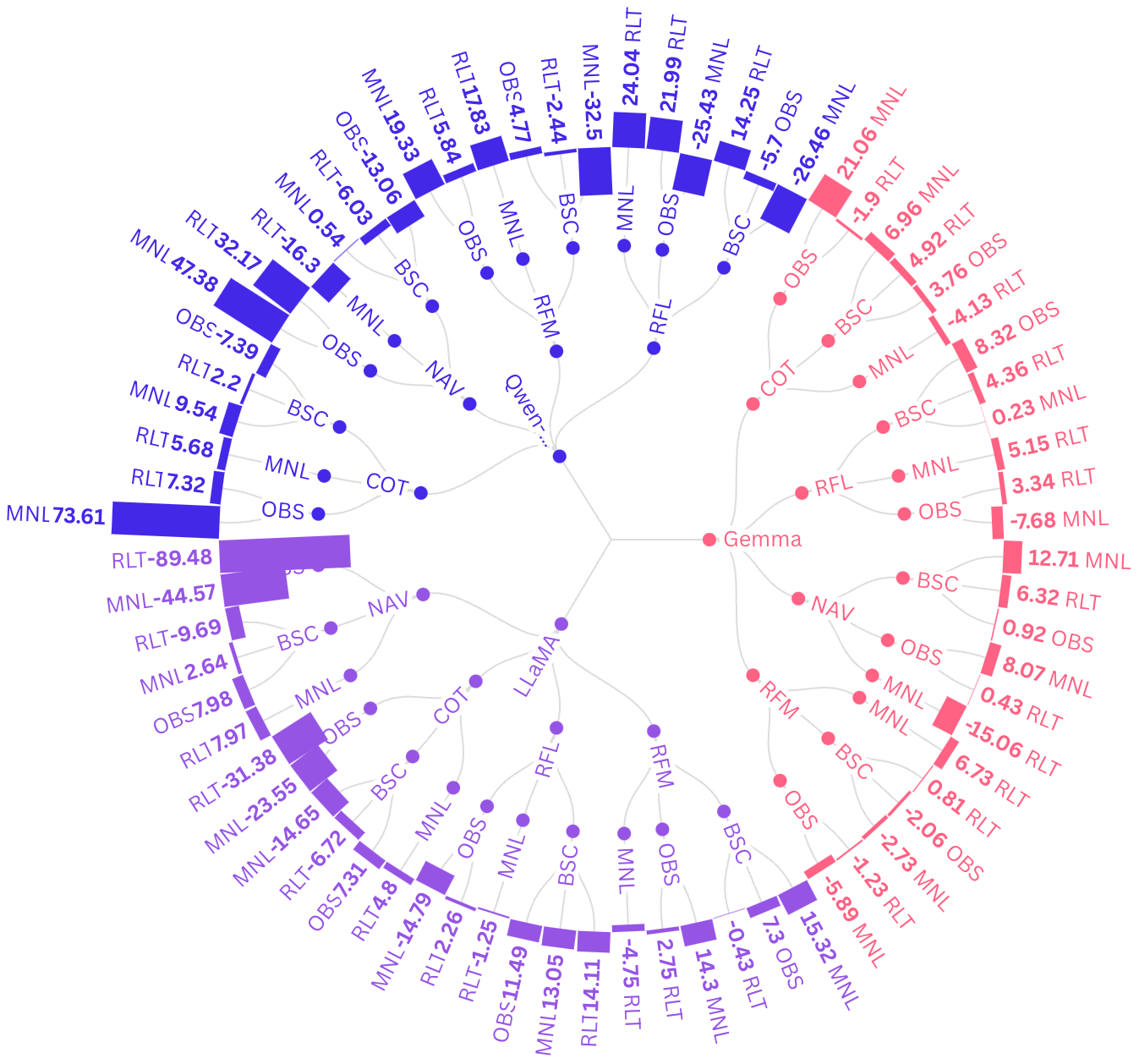}
        \label{fig:subfig1-both}
    \end{subfigure}
    \begin{subfigure}[b]{\linewidth}
        \centering
        \includegraphics[width=\linewidth]{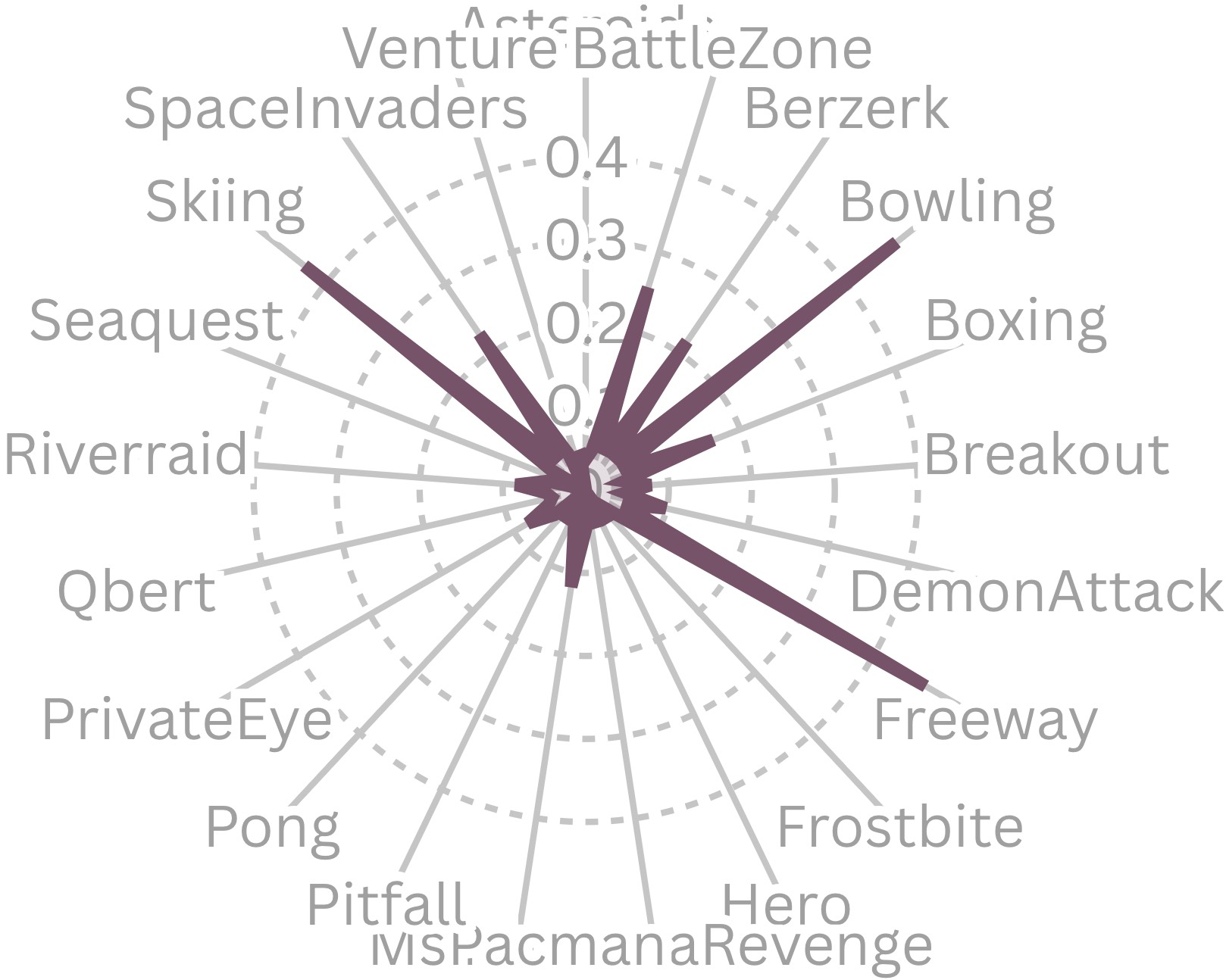}
        \label{fig:subfig2-both}
    \end{subfigure}
    \caption{Relative performances.}
    \label{fig:both}
\end{wrapfigure}

\vspace{0.1cm}
\noindent\textbf{Chain-of-Thought.}
The CoT agent extends the Basic template by demanding explicit step-by-step reasoning. A leading \texttt{system} instruction frames the model as an expert Atari player and mandates a JSON reply with two keys--\texttt{"thought process"} and \texttt{"action"}. Immediately thereafter the agent appends the game synopsis, the win or termination clause, and the latest observation, followed by a \textbf{fixed user instruction template} that drives the reasoning routine.
This prompt asks the model to pause, articulate its internal deliberation, and emit a machine-readable decision. To stabilise style, a handful of demonstration dialogues illustrating ideal chain-of-thought reasoning are prefixed; these exemplars remain fixed across games. Optional long-term memory (archived reasoning traces and episode rewards) and short-term memory (the most recent state--action pairs) are inserted when token budget permits. A running counter tracks prompt length and discards the oldest background messages--first long-term memory, then exemplars--once the predicted total nears the context window, ensuring a deliverable query at every step.

\vspace{0.1cm}
\noindent\textbf{CoT with Reflection.}
The Reflection agent carries out step-wise reasoning exactly as in CoT but inserts a self-critique phase at the end of every episode. When the terminal state is reached, the agent supplies the model with its complete thought trace, the realised reward trajectory, and the following system instruction.
The model's JSON reply is stored as a \textit{reflection block}. At the start of the next episode up to three of the most recent reflection blocks are inserted as additional \texttt{system} messages labelled \textit{Recent Plans}, giving the model a concise self-generated briefing on past mistakes, lessons, and intended adjustments. Older reflections are discarded whenever their inclusion would push the prompt beyond the context window, ensuring consistent token budgets. During gameplay the per-step prompt and action-parsing logic remain identical to CoT, but every episode begins with a short, data-driven feedback loop that encourages iterative policy refinement without parameter updates.

\subsection{Results}

\begin{figure}[htb!]
    \centering
    \setlength{\belowcaptionskip}{0pt}
    \setlength{\abovecaptionskip}{0pt}
    
    \begin{subfigure}[b]{\linewidth}
        \centering
        \includegraphics[width=\linewidth]{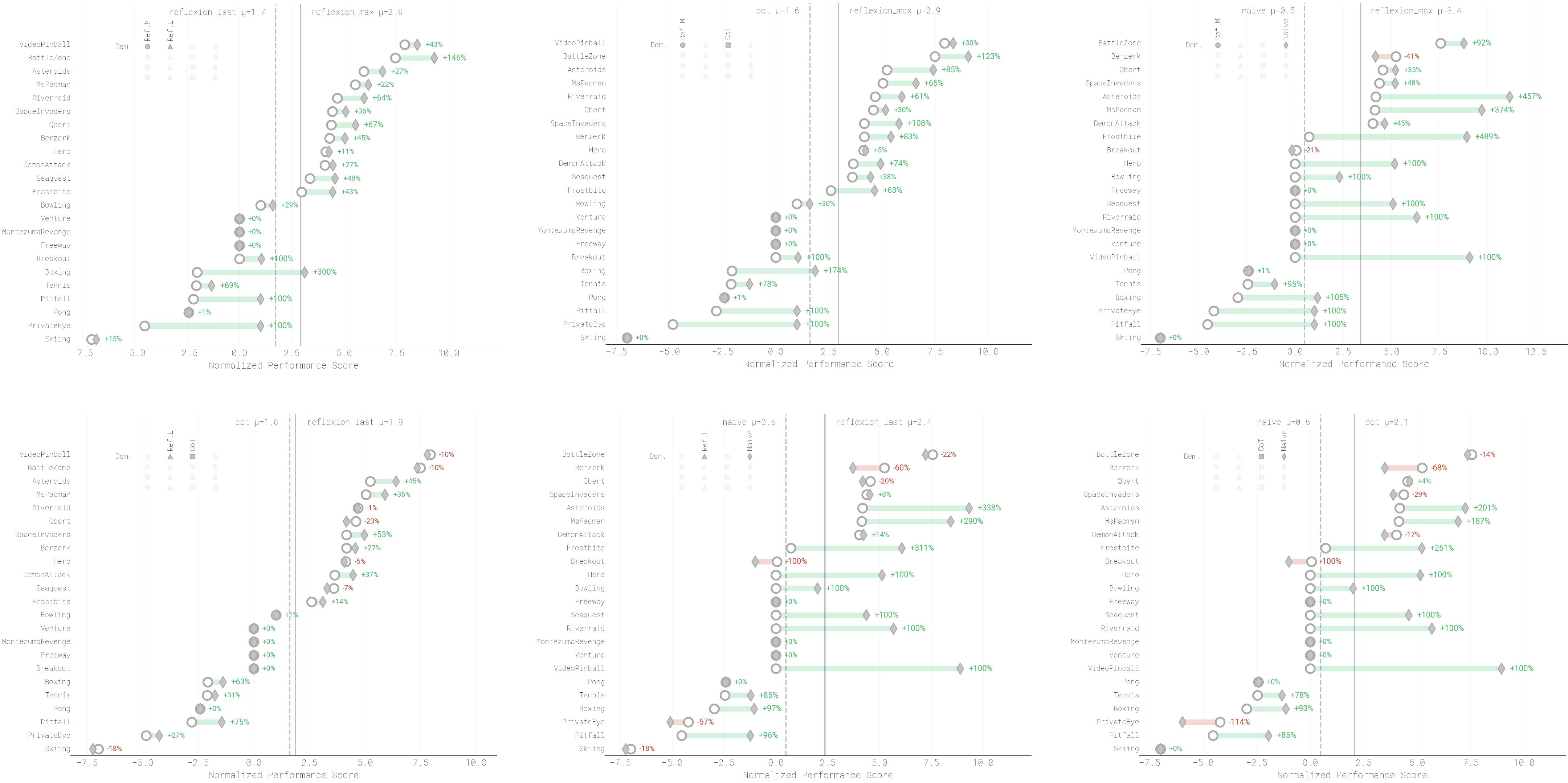}
        \label{fig:subfig1}
    \end{subfigure}
    
    \vspace{-0.5ex}
    
    \begin{subfigure}[b]{\linewidth}
        \centering
        \includegraphics[width=\linewidth]{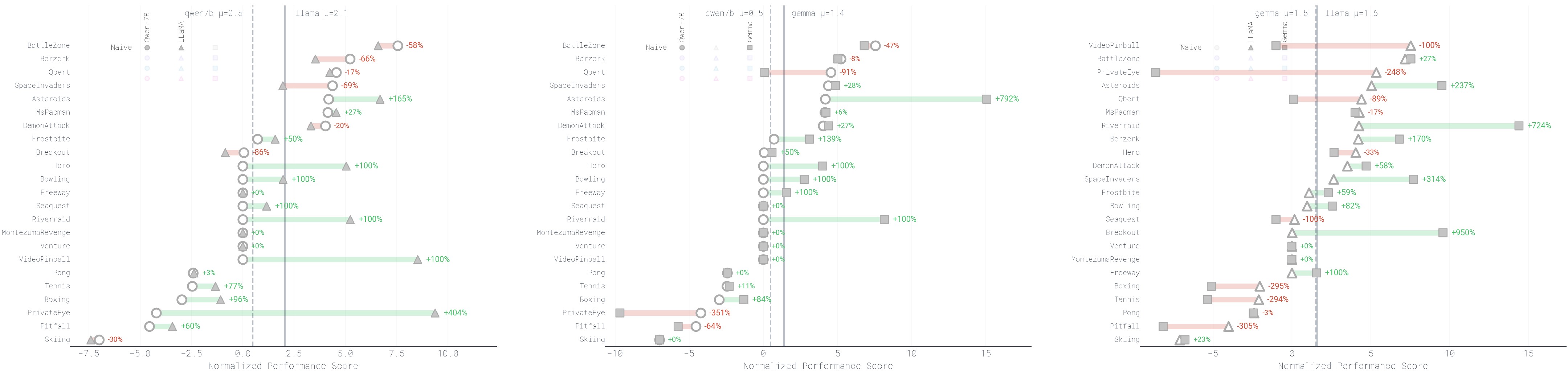}
        \label{fig:subfig2}
    \end{subfigure}
    
    \vspace{-0.5ex}
    
    \begin{subfigure}[b]{\linewidth}
        \centering
        \includegraphics[width=\linewidth]{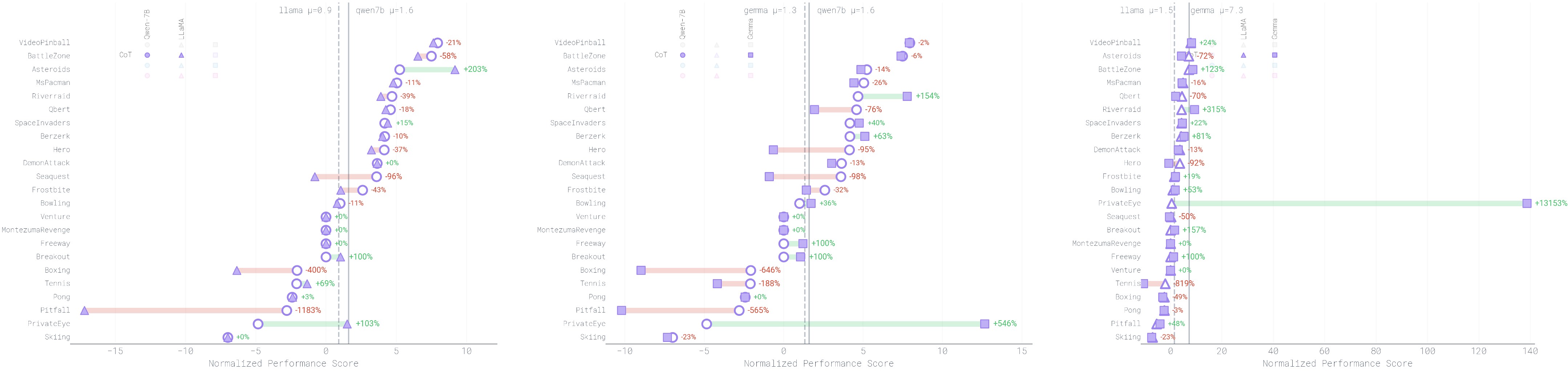}
        \label{fig:subfig3}
    \end{subfigure}
    
    \caption{Selected performance comparison. See Appendix for more details.}
    \label{fig:stacked}
    \vspace{-20pt}
\end{figure}

We evaluate three open-source large language models (Qwen2.5-7B, Gemma-7B, and Llama3.1-8B) across three agent frameworks (zero-shot, few-shot chain-of-thought, and reflection reasoning) on TextAtari. 
Our findings reveal that language models struggle significantly with very long-horizon tasks, with performance across over $90$\% of scenarios falling below $10$\% of human capability. 
Only in two specific tasks did the best agent configurations approach or marginally exceed human performance. 
Prior knowledge integration—specifically game manuals and expert demonstrations—emerged as the most consistent performance driver, yielding average improvements exceeding $100$\% across models and tasks. 
Surprisingly, reasoning-enhancement techniques such as chain-of-thought prompting and reflection mechanisms showed inconsistent benefits, with effectiveness varying dramatically across model architectures and game environments. 
This variability underscores the fundamental challenges in maintaining coherent state tracking, strategic planning, and decision consistency across extended time horizons. 
The substantial performance gap between even the best language agents and human players highlights the difficulty of maintaining coherent decision-making over tens of thousands of steps, suggesting that current language models, regardless of architecture or prompting technique, lack the cognitive mechanisms necessary for truly extended reasoning.

\section{Conclusion}\label{sec:conclusion}

We introduced TextAtari, a comprehensive benchmark for evaluating language agents on very long-horizon decision-making tasks spanning up to $100,000$ steps. By transforming visual Atari games into textual descriptions, we created a challenging testbed bridging sequential decision-making with natural language processing.
Our experimental results revealed significant challenges in long-horizon planning for current language models. Even the best agent configurations achieved less than $10$\% of human performance across over $90$\% of tasks, with only two specific scenarios approaching human-level competence. Prior knowledge integration (game manuals and expert demonstrations) yielded the most consistent improvements, averaging over $100$\% performance gains, while reasoning enhancement techniques showed inconsistent benefits across model architectures and environments.
TextAtari addresses a critical gap in existing benchmarks by specifically targeting language-based reasoning over extended temporal scales. Progress on this benchmark could advance autonomous agent development toward systems capable of maintaining consistent reasoning and planning at human-like scales.

\noindent\textbf{Limitations and future work} 
TextAtari has several limitations suggesting directions for future research. The benchmark's substantial computational demands—approximately $820,000$ GPU-minutes with some games requiring over $300$ hours per model and up to 50k tokens per decision step—may limit accessibility. Future work should explore more efficient evaluation protocols without sacrificing benchmark validity.
The transformation of visual environments into text, while methodologically sound, introduces potential information loss. Future iterations could explore multimodal extensions and evaluate how different textual representation granularities affect performance. Additionally, as new architectures emerge specifically targeting sequential reasoning and memory management, TextAtari should evolve accordingly.
Finally, while our benchmark demonstrates the significant gap between current AI systems and human capabilities, it does not fully diagnose the specific cognitive mechanisms responsible for this discrepancy. Future work could incorporate more detailed error analysis and causal interventions to identify specific reasoning bottlenecks, guiding targeted architectural improvements for long-horizon reasoning in language models.

\clearpage
\newpage

\bibliographystyle{icml2025}
\bibliography{main}


\newpage

\clearpage
\newpage

\appendix

\section{TextAtari Supplementary Material}
\begin{itemize}
    \item Section~\ref{sec:related-work}: Related Work
    \item Section~\ref{sec:task-details}: Task Details
    \item Section~\ref{sec:prompt}: Prompt Engineering
    \item Section~\ref{sec:missing-results}: Missing Results
    \item Section~\ref{sec:impact}: Border Impact
\end{itemize}

\section{Related Work}\label{sec:related-work}

This section provides a comprehensive analysis of existing sequential decision-making benchmarks to contextualize TextAtari's contribution. 
We surveyed 163 benchmarks across multiple domains—Video Games, Web Interactions, Software Operations, Text Games, Card Games, and Embodied Tasks—to evaluate the current landscape of agent evaluation frameworks.

Our analysis reveals a critical limitation in existing benchmarks: 
most operate on remarkably short horizons, with the median decision step count remaining below 50 for web and software manipulation tasks. 
Even among text and video games, which represent the longest horizon challenges, 75\% approach only 500 steps. 
As illustrated in Figure 3 of the main paper, fewer than 2.5\% of existing benchmarks reach the 100,000-step threshold that TextAtari addresses.

\begin{table}[htb!]
\centering
\caption{Comprehensive language agent benchmark collection for sequential decision making (Part I). This table catalogs benchmarks for evaluating language agents across diverse domains. Each benchmark is characterized by its domain category (e.g., Video Game, Web, Software, Text Game, Card Game, Embodied), decision horizon (number of steps required for task completion), number of distinct tasks, and agent type (single-agent or multi-agent). This collection contextualizes TextAtari's contribution to the benchmark landscape, particularly in addressing the challenge of very long-horizon decision-making tasks (up to 100,000 steps) for language agents.}
\vspace{5pt}
\label{tab:benchmarks-part1}
\resizebox{\textwidth}{!}{%
\begin{tabular}{clllcl}
\toprule
\textbf{ID} & \textbf{Name} & \textbf{Category} & \textbf{Horizon} & \textbf{Tasks} & \textbf{Agent Type} \\
\midrule
1 & StarCraft II~\citep{ma2024large} & Video Game & 1000 & 10 & multi-agent \\
2 & Red Dead Redemption II~\citepapp{tan2024towards} & Video Game & 1000 & 5 & single-agent \\
3 & V-MAGE~\citepapp{zheng2025v} & Video Game & 100, 1000, 5000 & 50 & single-agent \\
4 & ONUW~\citepapp{jin2024learning} & Card Game & 1 & 1 & multi-agent \\
5 & WebGames~\citepapp{thomas2025webgames} & Web & 50 & 50 & single-agent \\
6 & BEARCUBS~\citepapp{song2025bearcubs} & Web & 10 & 100 & single-agent \\
7 & BattleAgentBench~\citepapp{wang2024battleagentbench} & Video Game & 100 & 10 & multi-agent \\
8 & Collab-Overcooked~\citepapp{sun2025collab} & Video Game & 10, 100 & 50 & multi-agent \\
9 & SwarmBench~\citepapp{ruan2025benchmarking} & Video Game & 100 & 5 & multi-agent \\
10 & DSGBench~\citepapp{tang2025dsgbench} & Card Game, Video Game & 10, 100, 1000 & 5 & multi-agent \\
11 & PokéChamp~\citepapp{karten2025pok} & Video Game & 10, 100 & 1 & multi-agent \\
12 & Minedojo~\citep{fan2022minedojo} & Video Game & 1000 & 5000 & single-agent \\
13 & TextWorld~\citepapp{cote2019textworld} & Text Game & 1000 & 50 & single-agent \\
14 & AlfWorld~\citepapp{shridharalfworld} & Embodied & 50 & 5000 & single-agent \\
15 & BabyAI-Text~\citepapp{carta2023grounding} & Video Game & 100, 1000 & 50 & single-agent \\
16 & AgentBench~\citepapp{liu2024agentbench} & Card Game, Embodied, Software, Web & 10, 50 & 10 & single-agent \\
17 & GameTraversalBenchmark~\citepapp{nasirgametraversalbenchmark} & Video Game & 10, 100 & 100 & single-agent \\
18 & WorkArena++\citepapp{boisvert2024workarena++} & Web & 1, 10 & 500 & single-agent \\
19 & EMBODIED AGENT INTERFACE\citepapp{li2024embodied} & Embodied & 50 & 100 & single-agent \\
20 & VLABench~\citepapp{zhang2024vlabench} & Embodied & 500 & 100 & single-agent \\
21 & MindCraft~\citepapp{white2025collaborating} & Video Game & 10, 100 & 5000 & multi-agent \\
22 & Multi-Mission Tool Bench~\citepapp{yu2025multi} & Software, Web & 50 & 1000 & single-agent \\
23 & SPREADSHEETBENCH~\citepapp{maspreadsheetbench} & Software & 10 & 1000 & single-agent \\
24 & OSWorld~\citepapp{xie2024osworld} & Software & 10 & 500 & single-agent \\
25 & WebCanvas~\citepapp{pan2024webcanvas} & Web & 10 & 1000 & single-agent \\
26 & TUR[K]INGBENCH~\citepapp{xu2024tur} & Web & 50 & 100 & single-agent \\
27 & Windows Agent Arena~\citepapp{bonatti2024windows} & Software, Web & 10 & 100 & single-agent \\
28 & VisualAgentBench~\citepapp{liu2024visualagentbench} & Embodied, Software, Web & 10 & 1000 & single-agent \\
29 & OFFICEBENCH~\citepapp{wang2024officebench} & Software & 50 & 500 & single-agent \\
30 & WONDERBREAD~\citepapp{wornow2024wonderbread} & Web & 10 & 500 & single-agent \\
31 & EscapeBench~\citepapp{qian2024escapebench} & Text Game & 100, 1000, 500 & 500 & single-agent \\
32 & VisEscape~\citepapp{lim2025visescape} & Video Game & 100, 500 & 1000 & single-agent \\
33 & B-MoCA~\citepapp{lee2024benchmarking} & Software & 10 & 100 & single-agent \\
34 & Spider2-V~\citepapp{cao2024spider2} & Software & 10, 50 & 500 & single-agent \\
35 & ELT-Bench~\citepapp{jin2025elt} & Software & 10, 100, 50 & 100 & single-agent \\
36 & VideoGUI~\citepapp{lin2024videogui} & Software & 10, 50 & 500 & single-agent \\
37 & TaskBench~\citepapp{shen2024taskbench} & Software, Web & 10 & 20000 & single-agent \\
38 & AQA-Bench~\citepapp{yang2024aqa} & Text Game & 10, 50 & 10 & single-agent \\
39 & FamilyTool~\citepapp{wang2025familytool} & Software, Web & 1, 5 & 500 & single-agent \\
40 & MirrorAPI~\citepapp{guo2025stabletoolbench} & Software, Web & 1, 5 & 5000 & single-agent \\
41 & WhodunitBench~\citepapp{xie2024whodunitbench} & Card Game & 10, 50 & 50 & multi-agent \\
42 & GTA~\citepapp{wang2024gta} & Software, Web & 1, 5 & 500 & single-agent \\
43 & m\&m's~\citepapp{ma2024m} & Software, Web & 1, 5 & 5000 & single-agent \\
44 & DISCOVERYWORLD~\citepapp{jansen2024discoveryworld} & Video Game & 1000 & 50 & single-agent \\
45 & debug-gym~\citepapp{yuan2025debug} & Software & 50 & 2000 & single-agent \\
46 & HCAST~\citepapp{rein2025hcast} & Software & 5, 50 & 200 & single-agent \\
47 & AdaSociety~\citepapp{huangadasociety} & Video Game & 50 & 5 & multi-agent \\
48 & Mars~\citepapp{tang2024mars} & Video Game & 1000, 50 & 10 & single-agent \\
49 & AndroidControl~\citepapp{li2024effects} & Software & 5 & 15000 & single-agent \\
50 & A3~\citepapp{chai2025a3} & Software & 10, 5, 50 & 200 & single-agent \\
51 & ROBOTOUILLE~\citepapp{gonzalez2025robotouille} & Video Game & 10, 50 & 50 & multi-agent \\
52 & MindAgent~\citepapp{gong2024mindagent} & Video Game & 10, 5 & 10 & multi-agent \\
53 & PlanBench~\citepapp{valmeekam2023planbench} & Text Game & 10, 50 & 1000 & single-agent \\
54 & $\tau$-bench~\citepapp{yao2024tau} & Software, Web & 5, 50 & 200 & single-agent \\
55 & TheAgentCompany~\citepapp{xu2024theagentcompany} & Software, Web & 10, 50 & 200 & single-agent \\
\bottomrule
\end{tabular}}
\end{table}

\begin{table}[htb!]
\centering
\caption{Comprehensive language agent benchmark collection for sequential decision making (Part II). This table catalogs benchmarks for evaluating language agents across diverse domains. Each benchmark is characterized by its domain category (e.g., Video Game, Web, Software, Text Game, Card Game, Embodied), decision horizon (number of steps required for task completion), number of distinct tasks, and agent type (single-agent or multi-agent). This collection contextualizes TextAtari's contribution to the benchmark landscape, particularly in addressing the challenge of very long-horizon decision-making tasks (up to 100,000 steps) for language agents.}
\vspace{5pt}
\label{tab:benchmarks-part2}
\resizebox{\textwidth}{!}{%
\begin{tabular}{clllcl}
\toprule
\textbf{ID} & \textbf{Name} & \textbf{Category} & \textbf{Horizon} & \textbf{Tasks} & \textbf{Agent Type} \\
\midrule
56 & WebArena~\citepapp{zhouwebarena} & Web & 10, 50 & 1000 & single-agent \\
57 & WebShop~\citepapp{yao2022webshop} & Web & 5, 50 & 10000 & single-agent \\
58 & SHORTCUTSBENCH~\citepapp{shen2024shortcutsbench} & Software, Web & 10, 50 & 10000 & single-agent \\
59 & BALROG~\citepapp{paglieri2024balrog} & Text Game, Video Game & 10, 100000 & 10 & single-agent \\
60 & MLGym~\citepapp{nathani2025mlgym} & Software, Web & 50 & 10 & single-agent \\
61 & VGRP-Bench~\citepapp{ren2025vgrp} & Text Game & 10, 50 & 20 & single-agent \\
62 & INVESTORBENCH~\citepapp{li2024investorbench} & Software & 250, 50 & 5 & single-agent \\
63 & AndroidWorld~\citepapp{rawles2024androidworld} & Software & 5, 50 & 100 & single-agent \\
64 & MobileAgentBench~\citepapp{wang2024mobileagentbench} & Software & 10, 5 & 100 & single-agent \\
65 & SPA-Bench~\citepapp{chen2024spa} & Software & 5, 50 & 500 & single-agent \\
66 & PARTNR~\citepapp{chang2024partnr} & Embodied & 10, 100 & 100000 & multi-agent \\
67 & LVLM-Playground~\citepapp{wang2025large} & Video Game & 100, 5 & 20 & multi-agent \\
68 & GameArena~\citepapp{hu2024gamearena} & Text Game & 10, 5 & 20 & single-agent \\
69 & TEXTARENA~\citepapp{guertler2025textarena} & Text Game & 10, 500 & 100 & multi-agent \\
70 & MinePlanner~\citepapp{hill2023mineplanner} & Video Game & 100, 5 & 50 & single-agent \\
71 & GAMEBENCH~\citepapp{costarelli2024gamebench} & Text Game & 250 & 10 & single-agent \\
72 & GTBENCH~\citepapp{duangtbench} & Text Game & 50 & 10 & multi-agent \\
73 & ING-VP~\citepapp{zhang2024ing} & Text Game, Video Game & 50 & 300 & single-agent \\
74 & RoCoBench~\citepapp{mandi2024roco} & Embodied & 20, 5 & 10 & multi-agent \\
75 & VillagerAgent~\citepapp{dong2024villageragent} & Video Game & 500 & 200 & multi-agent \\
76 & LLMARENA~\citepapp{chen2024llmarena} & Text Game & 10, 200 & 10 & multi-agent \\
77 & CivRealm~\citepapp{qicivrealm} & Video Game & 1000 & 10 & multi-agent \\
78 & SmartPlay~\citepapp{wusmartplay} & Video Game & 100, 100000, 200, 5 & 10 & single-agent \\
79 & MAgIC~\citepapp{xu2024magic} & Card Game & 20, 5 & 5 & multi-agent \\
80 & AgentBoard~\citepapp{maagentboard} & Embodied, Software, Text Game, Web & 10, 20, 50 & 10 & single-agent \\
81 & AGENTGYM~\citepapp{xi2024agentgym} & Embodied, Software, Text Game, Web & 20 & 100 & single-agent \\
82 & Welfare Diplomacy~\citepapp{mukobiwelfare} & Card Game & 1, 10 & 1 & multi-agent \\
83 & Werewolf Arena~\citepapp{bailis2024werewolf} & Card Game & 10 & 5 & multi-agent \\
84 & MiniWoB~\citepapp{shi2017world} & Web & 1, 10 & 100 & single-agent \\
85 & MiniWoB++\citepapp{liu2018reinforcement} & Web & 1, 10 & 100 & single-agent \\
86 & WorkArena\citepapp{drouin2024workarena} & Web & 10, 20 & 100 & single-agent \\
87 & ManiSkill~\citepapp{mu2021maniskill} & Embodied & 50 & 20 & single-agent \\
88 & LIBERO~\citepapp{liu2023libero} & Embodied & 100 & 100 & single-agent \\
89 & RoboCasa~\citepapp{nasiriany2024robocasa} & Embodied & 500 & 100 & single-agent \\
90 & ARNOLD~\citepapp{gong2023arnold} & Embodied & 100 & 10 & single-agent \\
91 & Rlbench~\citepapp{james2020rlbench} & Embodied & 200 & 100 & single-agent \\
92 & Overcooked-AI~\citepapp{carroll2019utility} & Video Game & 500 & 5 & multi-agent \\
93 & CerealBar~\citepapp{perez2023measuring} & Video Game & 500 & 1 & multi-agent \\
94 & LLM-Coordination~\citepapp{agashe2023llm} & Video Game & 50, 500 & 5 & multi-agent \\
95 & MineLand~\citepapp{yu2024mineland} & Video Game & 2000 & 5000 & multi-agent \\
96 & APIBench~\citepapp{peng2022revisiting} & Software, Web & 1 & 20000 & single-agent \\
97 & ToolBench~\citepapp{xu2023tool} & Software, Web & 5 & 100000 & single-agent \\
98 & API-Bank~\citepapp{li2023api} & Software, Web & 2 & 200 & single-agent \\
99 & ToolAlpaca~\citepapp{tang2023toolalpaca} & Software, Web & 1 & 5000 & single-agent \\
100 & T-EVAL~\citepapp{chen2024t} & Software, Web & 5 & 20000 & single-agent \\
101 & UltraTool~\citepapp{huang2024planning} & Software, Web & 20 & 5000 & single-agent \\
102 & SheetCopilotBench~\citepapp{li2023sheetcopilot} & Software & 10 & 200 & single-agent \\
103 & InstructExcel~\citepapp{payan2023instructexcel} & Software & 10 & 5000 & single-agent \\
104 & SheetRM~\citepapp{chen2024sheetagent} & Software & 10 & 500 & single-agent \\
105 & GAIA~\citepapp{mialon2023gaia} & Software, Web & 50 & 500 & single-agent \\
106 & Mind2Web~\citepapp{deng2023mind2web} & Web & 10 & 2000 & single-agent \\
107 & WEBLINX~\citepapp{lu2024weblinx} & Web & 50 & 2000 & single-agent \\
108 & METAGUI~\citepapp{sun2022meta} & Software & 5 & 1000 & single-agent \\
109 & AITW~\citepapp{rawles2307android} & Software & 5 & 30000 & single-agent \\
110 & PIXELHELP~\citepapp{li2020mapping} & Software & 5 & 200 & single-agent \\
\bottomrule
\end{tabular}}
\end{table}

\begin{table}[htb!]
\centering
\caption{Comprehensive language agent benchmark collection for sequential decision making (Part III). This table catalogs benchmarks for evaluating language agents across diverse domains. Each benchmark is characterized by its domain category (e.g., Video Game, Web, Software, Text Game, Card Game, Embodied), decision horizon (number of steps required for task completion), number of distinct tasks, and agent type (single-agent or multi-agent). This collection contextualizes TextAtari's contribution to the benchmark landscape, particularly in addressing the challenge of very long-horizon decision-making tasks (up to 100,000 steps) for language agents.}
\vspace{5pt}
\label{tab:benchmarks-part3}
\resizebox{\textwidth}{!}{%
\begin{tabular}{clllcl}
\toprule
\textbf{ID} & \textbf{Name} & \textbf{Category} & \textbf{Horizon} & \textbf{Tasks} & \textbf{Agent Type} \\
\midrule
111 & OmniAct~\citepapp{kapoor2024omniact} & Software, Web & 5 & 10000 & single-agent \\
112 & InterCode~\citepapp{yang2023intercode} & Software & 10 & 1000 & single-agent \\
113 & VisualWebArena~\citepapp{koh2024visualwebarena} & Web & 50 & 1000 & single-agent \\
114 & Mobile-Env~\citepapp{zhang2023mobile} & Software, Web & 10 & 200 & single-agent \\
115 & AssistGUI~\citepapp{gao2023assistgui} & Software & 5 & 100 & single-agent \\
116 & ScienceWorld~\citepapp{wang2022scienceworld} & Video Game & 100 & 50 & single-agent \\
117 & MoTIF~\citepapp{klissarovmotif} & Software & 20 & 5000 & single-agent \\
118 & MLAgentBench~\citepapp{huang2024mlagentbench} & Software & 20 & 10 & single-agent \\
119 & Spider 2.0~\citepapp{leispider} & Software & 50 & 500 & single-agent \\
120 & MetaTool~\citepapp{wang2024metatool} & Software, Web & 5 & 20000 & single-agent \\
121 & DDD~\citepapp{wu2024deciphering} & Card Game & 10 & 5 & multi-agent \\
122 & AvalonBench~\citepapp{light2023avalonbench} & Card Game & 20 & 1 & multi-agent \\
123 & SOTOPIA~\citepapp{zhousotopia} & Text Game & 20 & 500 & multi-agent \\
124 & ToolEmu~\citepapp{ruanidentifying} & Software & 10 & 100 & single-agent \\
125 & MiniGrid~\citepapp{chevalier2023minigrid} & Video Game & 100 & 20 & single-agent \\
126 & Alfred~\citepapp{shridhar2020alfred} & Embodied & 50 & 5 & single-agent \\
127 & NetHack LE~\citepapp{kuttler2020nethack} & Text Game & 100000 & 10 & single-agent \\
128 & Alchemy~\citepapp{chen2019alchemy} & Video Game & 200 & 1 & single-agent \\
129 & IVRE~\citepapp{xu2023interactive} & Video Game & 10 & 1 & single-agent \\
130 & UGIF~\citepapp{venkatesh2022ugif} & Software & 5 & 500 & single-agent \\
131 & WebVoyager~\citepapp{he2024webvoyager} & Web & 500 & 20 & single-agent \\
132 & AMEX~\citepapp{chai2024amex} & Software & 10 & 3000 & single-agent \\
133 & AndroidArena~\citepapp{xing2024understanding} & Software & 10 & 200 & single-agent \\
134 & AndroidLab~\citepapp{xu2024androidlab} & Software & 30 & 100 & single-agent \\
135 & ARA~\citepapp{alghamdi2024aratrust} & Software & 5 & 10 & single-agent \\
136 & AsyncHow~\citepapp{lin2024graph} & Text Game & 10 & 2000 & single-agent \\
137 & VirtualHome~\citepapp{puig2018virtualhome} & Embodied & 100 & 3000 & single-agent \\
138 & WAH~\citepapp{puig2020watch} & Embodied & 250 & 1000 & multi-agent \\
139 & VideoWebArena~\citepapp{jang2024videowebarena} & Web & 20 & 2000 & single-agent \\
140 & HandMeThat~\citepapp{wan2022handmethat} & Embodied & 50 & 30000 & multi-agent \\
141 & DialFRED~\citepapp{gao2022dialfred} & Embodied & 50 & 30000 & multi-agent \\
142 & TEACh~\citepapp{padmakumar2022teach} & Embodied & 100 & 5000 & multi-agent \\
143 & LIGHT~\citepapp{urbanek2019learning} & Text Game & 200 & 10000 & multi-agent \\
144 & Diplomacy~\citepapp{bakhtin2021no} & Card Game & 100 & 1 & multi-agent \\
145 & AppWorld~\citepapp{trivedi2024appworld} & Software & 20 & 1000 & single-agent \\
146 & ToolLLM~\citepapp{qintoolllm} & Software, Web & 5 & 10000 & single-agent \\
148 & ToolQA~\citepapp{zhuang2023toolqa} & Software, Web & 5 & 2000 & single-agent \\
149 & ToolLens~\citepapp{qu2024towards} & Software, Web & 5 & 20000 & single-agent \\
150 & Crafter~\citepapp{hafnerbenchmarking} & Video Game & 1000 & 20 & single-agent \\
151 & Baba is AI~\citepapp{cloos2024baba} & Video Game & 100 & 5 & single-agent \\
152 & MiniHack~\citepapp{samvelyan1minihack} & Text Game & 100 & 100 & single-agent \\
153 & MLE-Bench~\citepapp{chan2024mle} & Software & 2000 & 100 & single-agent \\
154 & RE-Bench~\citepapp{wijk2024re} & Software & 5000 & 10 & single-agent \\
155 & ScienceAgentBench~\citepapp{chen2024scienceagentbench} & Software & 10 & 100 & single-agent \\
156 & LlamaTouch~\citepapp{zhang2024llamatouch} & Software & 50 & 500 & single-agent \\
157 & AgentStudio~\citepapp{zheng2024agentstudio} & Software & 10 & 200 & single-agent \\
158 & RoboGen~\citepapp{wang2024robogen} & Embodied & 10 & 100 & single-agent \\
159 & Clembench~\citepapp{chalamalasetti2023clembench} & Text Game & 10 & 200 & multi-agent \\
160 & LMRL-Gym~\citepapp{abdulhai2023lmrl} & Text Game & 100 & 10 & multi-agent \\
161 & Game-theoretic LLM~\citepapp{hua2024game} & Text Game & 20 & 10 & multi-agent \\
162 & LAMEN~\citepapp{davidson2024evaluating} & Text Game & 10 & 5 & multi-agent \\
163 & SPIN-Bench~\citepapp{yao2025spin} & Text Game & 50 & 5 & multi-agent \\
\bottomrule
\end{tabular}}
\end{table}

Table~\ref{tab:benchmarks-part1}-\ref{tab:benchmarks-part3} catalog these benchmarks, highlighting key characteristics including domain category, decision horizon (number of steps required for task completion), number of distinct tasks, and agent type (single-agent or multi-agent). 
This collection provides empirical evidence for the ``horizon gap'' discussed in the introduction—the absence of standardized evaluations for truly long-horizon sequential decision-making tasks that would take humans days or weeks to complete.

TextAtari addresses this gap by providing a standardized framework for evaluating language agents' capacity to maintain coherent reasoning and decision-making across extended horizons of up to 100,000 steps. 
Unlike most existing benchmarks that focus on short-term tactical decisions, TextAtari challenges agents with long-term strategic planning where consequences compound over tens of thousands of steps—a qualitative shift in reasoning requirements that better approximates real-world extended planning challenges.

\subsection{Compare to Existing 100K-Step Benchmarks}

NetHack Learning Environment~\citepapp{kuttler2020nethack, paglieri2024balrog}, while technically supporting long gameplay sessions of tens of thousands of steps like TextAtari, suffers from several critical limitations. 
The complexity of NetHack creates an extremely sparse reward landscape that makes systematic evaluation challenging—agents often fail catastrophically before meaningful learning can occur. 
Additionally, NetHack's representation combines ASCII symbols with complex game mechanics that require substantial domain expertise to interpret properly. 
This creates an artificial hurdle of game-specific knowledge rather than testing general reasoning capabilities. 
Unlike TextAtari, which provides rich natural language descriptions accessible to any language model, NetHack's symbolic representation obscures the underlying state, making it difficult to disentangle reasoning failures from representation understanding issues.

SmartPlay~\citepapp{wusmartplay}, while built on Minecraft and theoretically supporting extended gameplay, fundamentally compromises on measuring true long-horizon planning by relying heavily on predefined macro actions. 
These high-level abstractions dramatically reduce the actual decision space—agents make far fewer genuine decisions than the step count suggests. 
This abstraction masks the fundamental challenges of maintaining coherent reasoning across extended sequences. 
In contrast, TextAtari preserves the raw granularity of decision-making, requiring agents to make individual primitive actions that compound over tens of thousands of steps, thus providing a more genuine measure of extended reasoning capabilities without artificial simplifications.

\subsection{Compare to Existing 1K-Step Benchmarks}

Many existing benchmarks that approach the thousand-step range suffer from various simplifications that reduce their effectiveness for evaluating true long-horizon reasoning. 
Game-based benchmarks like StarCraft II~\citep{ma2024large}, Red Dead Redemption II~\citepapp{tan2024towards}, and Minecraft variants (MineDojo~\citep{fan2022minedojo}, Mars~\citepapp{tang2024mars}, MineLand~\citepapp{yu2024mineland}, Crafter~\citepapp{hafnerbenchmarking}) all rely on predefined macro-actions or action abstractions that dramatically simplify the actual planning required. 
These abstract actions encapsulate complex sequences of decisions into single steps, creating an illusion of long-horizon planning while actually testing much shorter decision sequences.

V-MAGE~\citepapp{zheng2025v} remains limited to just two specific environments (SuperMario and FlappyBird), providing insufficient diversity to evaluate general reasoning capabilities across different domains. 
Its narrow focus on platforming mechanics fails to test the breadth of reasoning types (strategic planning, resource management, spatial understanding) that TextAtari's diverse game selection enables.

Text-based environments like TextWorld~\citepapp{cote2019textworld} and EscapeBench~\citepapp{qian2024escapebench} offer simplified worlds with limited state spaces and highly constrained action possibilities. 
Their deliberately simplified environments lack the complexity, state space size, and causal depth of Atari games. 
BabyAI-Text~\citepapp{carta2023grounding} similarly restricts itself to basic grid-world scenarios with limited objects and interactions, creating artificially simplified planning problems.

Benchmarks like MLE-Bench~\citepapp{chan2024mle}, RE-Bench~\citepapp{wijk2024re}, and DISCOVERYWORLD~\citepapp{jansen2024discoveryworld} limit themselves to specialized domains (machine learning experiments and scientific discovery) that contain substantial human annotation and guidance. 
These embedded hints and structured exploration spaces implicitly simplify the planning challenge compared to TextAtari's more general game environments where agents must discover effective strategies independently.

CivRealm~\citepapp{qicivrealm}, while offering complex strategic gameplay, suffers from an extremely restricted action space at each decision point combined with extensive domain-specific knowledge requirements. 
The game's built-in advisors and infrastructure for managing civilization development significantly reduce the actual reasoning burden on agents. 
Moreover, its specialized domain of civilization building lacks the diversity of reasoning types required by TextAtari's varied environments.

TextAtari overcomes these limitations by providing: 
(1) diverse environments spanning multiple reasoning types without domain specialization; 
(2) primitive action spaces that require genuine sequential decision-making without macro-action shortcuts; 
(3) natural language descriptions that eliminate the need for specialized visual processing while preserving state information; 
(4) standardized evaluation protocols across games with varying difficulty levels; and 
(5) true long-horizon challenges where decisions have compounding consequences over tens of thousands of steps. 
These advantages make TextAtari uniquely positioned to evaluate language agents' capacity for extended reasoning in a way that existing benchmarks cannot match.

\subsection{Future Direction}

Despite TextAtari's strengths in evaluating long-horizon reasoning, we acknowledge that existing benchmarks offer valuable perspectives our work can benefit from. 
Complex video games like StarCraft II and Red Dead Redemption II present rich, visually grounded environments with emergent dynamics that more closely mirror real-world complexity. 
While their macro-action approach simplifies decision horizons, these games capture strategic depth and environmental richness that complement TextAtari's focus on extended reasoning. 
Similarly, realistic task environments like DISCOVERYWORLD and MLE-Bench provide domain-specific challenges that better represent specialized professional reasoning in scientific discovery and machine learning experimentation—contexts where language agents may ultimately provide significant value.

We view TextAtari not as a replacement for these specialized benchmarks but as addressing a specific gap in evaluating \textbf{pure long-horizon reasoning capacity}. 
Our future work aims to develop a more comprehensive \textit{meta-benchmark} that integrates these complementary strengths—combining TextAtari's extended planning horizons with the visual grounding of complex video games, the specialized reasoning of professional domains, and the embodied interaction of Minecraft-style environments. 
This integrated approach would enable more holistic evaluation of language agents across multiple dimensions of intelligence: 
sustained reasoning over time, multimodal understanding, specialized domain knowledge application, and adaptive real-time control—potentially revealing capabilities and limitations that no single benchmark could identify in isolation.

\section{More Task Details}\label{sec:task-details}

This appendix provides comprehensive information about the Atari games used in our TextAtari benchmark. 
The following tables present detailed classifications and descriptions of all 23 Atari games included in our evaluation, organized by game category with specific challenges they pose for language models.

Tables~\ref{table:atari_games_appendix1} and~\ref{table:atari_games_appendix2} classify the games into four major categories (Action Games, Puzzle and Strategy Games, Sports Games, and Arcade Classics) and highlight the specific cognitive challenges each game presents for language models. 
These challenges are color-coded to indicate their primary nature: 
spatial reasoning, planning and strategy, partial observability, and temporal reasoning. 
This classification helps illuminate why certain games prove more difficult for language agents than others, and which cognitive capabilities are most critical for success across different game environments.

\begin{table}[htb!]
\begin{adjustbox}{center, width=\textwidth}
\small
\begin{tabularx}{\textwidth}{>{\hsize=0.4\hsize}X>{\hsize=0.6\hsize}X>{\hsize=1.0\hsize}X}
\rowcolor{header} \multicolumn{3}{c}{\textbf{Atari Games Classification and LLM Gaming Challenges (Part 1)}} \\
\midrule
\textbf{Game} & \textbf{Category} & \textbf{Challenges for LLM} \\
\midrule
\rowcolor{gray!10} \multicolumn{3}{c}{\textbf{Action Games}} \\
\midrule
Asteroids & Space Shooter & 
\begin{tabular}{p{0.9\hsize}}
\cellcolor{spatial}Spatial reasoning for circular topology \\
\cellcolor{temporal}Reaction-based gameplay timing \\
\cellcolor{planning}Continuous state space navigation \\
\cellcolor{planning}Strategic target prioritization
\end{tabular} \\
\midrule
BattleZone & First-Person Tank & 
\begin{tabular}{p{0.9\hsize}}
\cellcolor{spatial}3D spatial reasoning \\
\cellcolor{partial}First-person partially observed environment \\
\cellcolor{planning}Strategic target selection \\
\cellcolor{planning}Situational awareness maintenance
\end{tabular} \\
\midrule
Berzerk & Maze Shooter & 
\begin{tabular}{p{0.9\hsize}}
\cellcolor{spatial}Navigating complex maze layouts \\
\cellcolor{partial}Dynamic obstacle avoidance \\
\cellcolor{temporal}Time pressure (Evil Otto pursuit) \\
\cellcolor{planning}Multitasking (walls, enemies, bullets)
\end{tabular} \\
\midrule
DemonAttack & Space Shooter & 
\begin{tabular}{p{0.9\hsize}}
\cellcolor{planning}Pattern recognition in enemy waves \\
\cellcolor{planning}Prediction of enemy movement patterns \\
\cellcolor{planning}Progressive difficulty adaptation \\
\cellcolor{temporal}Timing of defensive movements
\end{tabular} \\
\midrule
Hero & Exploration & 
\begin{tabular}{p{0.9\hsize}}
\cellcolor{planning}Resource management (dynamite) \\
\cellcolor{spatial}Complex navigation in caverns \\
\cellcolor{planning}Long-term planning for rescue \\
\cellcolor{partial}Non-linear exploration paths
\end{tabular} \\
\midrule
Pitfall & Platformer & 
\begin{tabular}{p{0.9\hsize}}
\cellcolor{temporal}Precise timing for obstacles \\
\cellcolor{spatial}Complex movement patterns \\
\cellcolor{planning}Long horizon optimization (20-minute gameplay) \\
\cellcolor{planning}Risk/reward assessment for treasure collection
\end{tabular} \\
\midrule
\rowcolor{gray!10} \multicolumn{3}{c}{\textbf{Puzzle and Strategy Games}} \\
\midrule
Breakout & Brick-breaker & 
\begin{tabular}{p{0.9\hsize}}
\cellcolor{spatial}Geometry understanding \\
\cellcolor{planning}Ball trajectory prediction \\
\cellcolor{planning}Strategic brick targeting \\
\cellcolor{temporal}Timing-sensitive paddle control
\end{tabular} \\
\midrule
Frostbite & Building & 
\begin{tabular}{p{0.9\hsize}}
\cellcolor{planning}Planning sequences under time constraints \\
\cellcolor{planning}Risk assessment with moving platforms \\
\cellcolor{planning}Multi-objective optimization (ice blocks vs. safety) \\
\cellcolor{temporal}Timing jumps between ice floes
\end{tabular} \\
\midrule
MontezumaRevenge & Puzzle Platformer & 
\begin{tabular}{p{0.9\hsize}}
\cellcolor{planning}Extremely sparse rewards \\
\cellcolor{planning}Long causal chains \\
\cellcolor{planning}Complex dependency hierarchies \\
\cellcolor{temporal}Precise timing for trap avoidance
\end{tabular} \\
\midrule
PrivateEye & Detective & 
\begin{tabular}{p{0.9\hsize}}
\cellcolor{planning}Sparse reward structure \\
\cellcolor{planning}Long-term memory requirements \\
\cellcolor{partial}Non-linear gameplay \\
\cellcolor{planning}Large state space navigation
\end{tabular} \\
\midrule
Qbert & Puzzle & 
\begin{tabular}{p{0.9\hsize}}
\cellcolor{spatial}Isometric spatial reasoning \\
\cellcolor{planning}Planning efficient color-changing pathways \\
\cellcolor{planning}Trap and enemy avoidance \\
\cellcolor{planning}Progressive difficulty adaptation
\end{tabular} \\
\bottomrule
\end{tabularx}
\end{adjustbox}
\caption{
\textbf{Detailed Analysis of Atari Games and Their LLM Challenges (Part 1).} This table presents action and puzzle games with their specific challenges for LLMs. Color coding indicates challenge types: \colorbox{spatial}{spatial reasoning}, \colorbox{planning}{planning \& strategy}, \colorbox{partial}{partial observability}, and \colorbox{temporal}{temporal reasoning}.
}
\label{table:atari_games_appendix1}
\end{table}

\begin{table}[htb!]
\begin{adjustbox}{center, width=\textwidth}
\small
\begin{tabularx}{\textwidth}{>{\hsize=0.4\hsize}X>{\hsize=0.6\hsize}X>{\hsize=1.0\hsize}X}
\rowcolor{header} \multicolumn{3}{c}{\textbf{Atari Games Classification and LLM Gaming Challenges (Part 2)}} \\
\midrule
\textbf{Game} & \textbf{Category} & \textbf{Challenges for LLM} \\
\midrule
\rowcolor{gray!10} \multicolumn{3}{c}{\textbf{Sports Games}} \\
\midrule
Bowling & Sports & 
\begin{tabular}{p{0.9\hsize}}
\cellcolor{spatial}Physics understanding \\
\cellcolor{planning}Precise parameter control \\
\cellcolor{planning}Adapting to pin configurations \\
\cellcolor{planning}Optimizing for strike probability
\end{tabular} \\
\midrule
Boxing & Sports & 
\begin{tabular}{p{0.9\hsize}}
\cellcolor{partial}Adversarial reasoning \\
\cellcolor{planning}Predicting opponent movements \\
\cellcolor{planning}Tactical positioning \\
\cellcolor{temporal}Timing attack and defense moves
\end{tabular} \\
\midrule
Pong & Sports & 
\begin{tabular}{p{0.9\hsize}}
\cellcolor{planning}Continuous control optimization \\
\cellcolor{planning}Anticipating ball trajectory \\
\cellcolor{temporal}Timing paddle movements \\
\cellcolor{spatial}Optimizing paddle positioning
\end{tabular} \\
\midrule
Skiing & Sports & 
\begin{tabular}{p{0.9\hsize}}
\cellcolor{temporal}Precise timing for gates \\
\cellcolor{planning}Path planning with fixed obstacles \\
\cellcolor{planning}Speed/accuracy trade-offs \\
\cellcolor{planning}Long-term performance optimization
\end{tabular} \\
\midrule
Tennis & Sports & 
\begin{tabular}{p{0.9\hsize}}
\cellcolor{spatial}Positioning for shots and court coverage \\
\cellcolor{partial}Anticipating opponent strategy \\
\cellcolor{planning}Shot selection and placement \\
\cellcolor{planning}Balancing offensive and defensive play
\end{tabular} \\
\midrule
\rowcolor{gray!10} \multicolumn{3}{c}{\textbf{Arcade Classics}} \\
\midrule
Freeway & Obstacle Avoidance & 
\begin{tabular}{p{0.9\hsize}}
\cellcolor{temporal}Timing road crossings \\
\cellcolor{planning}Pattern recognition in traffic flow \\
\cellcolor{planning}Binary sparse reward navigation \\
\cellcolor{planning}Risk assessment under time pressure
\end{tabular} \\
\midrule
MsPacman & Maze & 
\begin{tabular}{p{0.9\hsize}}
\cellcolor{planning}Dynamic path planning \\
\cellcolor{partial}Ghost behavior modeling (different behaviors) \\
\cellcolor{planning}Strategic power pellet usage \\
\cellcolor{planning}Adapting to maze layout variations
\end{tabular} \\
\midrule
Riverraid & Scrolling Shooter & 
\begin{tabular}{p{0.9\hsize}}
\cellcolor{planning}Resource management (fuel) \\
\cellcolor{planning}Prioritizing various hazard types \\
\cellcolor{planning}Adapting to increasing difficulty \\
\cellcolor{spatial}Navigating narrow passages
\end{tabular} \\
\midrule
Seaquest & Underwater Shooter & 
\begin{tabular}{p{0.9\hsize}}
\cellcolor{planning}Resource management (oxygen) \\
\cellcolor{planning}Multi-objective balancing (rescue, combat, survival) \\
\cellcolor{partial}Bidirectional threat assessment \\
\cellcolor{planning}Risk/reward decisions for surfacing
\end{tabular} \\
\midrule
SpaceInvaders & Space Shooter & 
\begin{tabular}{p{0.9\hsize}}
\cellcolor{planning}Strategy shifts based on remaining enemies \\
\cellcolor{planning}Managing defensive shelters \\
\cellcolor{planning}Adapting to increasing enemy speed \\
\cellcolor{spatial}Spatial awareness for shelter usage
\end{tabular} \\
\midrule
Venture & Dungeon Crawler & 
\begin{tabular}{p{0.9\hsize}}
\cellcolor{planning}Room-to-room navigation strategy \\
\cellcolor{partial}Partially observed state \\
\cellcolor{planning}Risk/reward assessment for treasures \\
\cellcolor{planning}Enemy pattern recognition
\end{tabular} \\
\midrule
VideoPinball & Simulation & 
\begin{tabular}{p{0.9\hsize}}
\cellcolor{spatial}Physics prediction \\
\cellcolor{temporal}Timing-based flipper control \\
\cellcolor{planning}Understanding complex scoring mechanisms \\
\cellcolor{planning}Long-term strategy for high scores
\end{tabular} \\
\bottomrule
\end{tabularx}
\end{adjustbox}
\caption{
\textbf{Detailed Analysis of Atari Games and Their LLM Challenges (Part 2).} This table presents sports games and arcade classics with their specific challenges for LLMs. Color coding indicates challenge types: \colorbox{spatial}{spatial reasoning}, \colorbox{planning}{planning \& strategy}, \colorbox{partial}{partial observability}, and \colorbox{temporal}{temporal reasoning}.
}
\label{table:atari_games_appendix2}
\end{table}

Tables~\ref{table:atari_games_description1}-\ref{table:atari_games_description4} provide comprehensive descriptions of each game's mechanics, objectives, and distinctive features. 
These detailed descriptions offer context for understanding the complexity of each environment and the specific demands they place on decision-making agents. 
The descriptions cover aspects such as control mechanisms, scoring systems, hazards, progression structures, and unique gameplay elements that distinguish each title.

\begin{table}[!t]
\begin{adjustbox}{center, width=\textwidth}
\small
\begin{tabularx}{\textwidth}{>{\hsize=0.27\hsize}X>{\hsize=0.73\hsize}X}
\rowcolor{header} \multicolumn{2}{c}{\textbf{Detailed Description of Atari Games (Part 1)}} \\
\midrule
\textbf{Game} & \textbf{Detailed Description} \\
\midrule
\rowcolor{lightrow} Asteroids & A space-themed shooter in vector graphics where the player controls a triangular ship in an asteroid field. The player must shoot and destroy asteroids while avoiding collisions. As asteroids are destroyed, they break into smaller pieces, creating more hazards. Occasionally, flying saucers appear and shoot at the player. The ship has momentum in a zero-gravity environment, requiring strategic thruster control and rotation. Players can hyperspace to a random location when in danger, though this carries risk. \\
\midrule
BattleZone & One of the first 3D tank combat games using vector graphics to create a first-person perspective. Players control a tank on a flat plain with mountains in the background and various obstacles like blocks and pyramids. Enemy tanks, missiles, and flying saucers attack the player, requiring strategic positioning and aiming. A radar display helps locate enemies outside the visible field. The realistic control scheme requires separate controls for driving and turret rotation, demanding sophisticated spatial coordination. \\
\midrule
\rowcolor{lightrow} Berzerk & A multi-directional shooter set in a maze of interconnected rooms. The player navigates through rooms filled with robots that shoot at the player. Touching robots, their bullets, or the electrified walls results in death. After a short time in any room, "Evil Otto" - an indestructible bouncing smiley face - appears and pursues the player, forcing quick movement to the next room. The robots' speech synthesis ("The humanoid must not escape!") was groundbreaking for its time. Each maze is procedurally generated, creating effectively endless gameplay. \\
\midrule
Bowling & A simulation of ten-pin bowling where players control the position and curvature of the ball's path. The player can position their bowler horizontally, set the ball's curve, and time the release for optimal accuracy. The game accurately models pin physics, with realistic pin scatter and knockdown patterns. Players compete across 10 frames, aiming for strikes and spares to maximize their score. Different pin configurations after the first throw require adaptive targeting strategies for picking up spares. \\
\midrule
\rowcolor{lightrow} Boxing & A top-down boxing simulation where two boxers face off in a ring. The player controls a boxer trying to land punches on the opponent while avoiding being hit. Players need to strategically position themselves, time their punches, and guard against counterattacks. The game has a three-minute time limit, and the winner is determined by either knockout or points scored through successful hits. Different punching angles and positions result in varying effectiveness, requiring strategic positioning and timing. \\
\midrule
Breakout & A brick-breaking game where the player controls a paddle at the bottom of the screen to bounce a ball upward into layers of bricks. When hit, bricks disappear, and the ball bounces back. The goal is to eliminate all bricks without letting the ball pass the paddle. As more bricks are destroyed, the ball moves faster. Some versions feature multi-colored brick layers with higher-value bricks at the top. Breaking through to the top allows the ball to bounce around the top edge, potentially clearing many bricks rapidly. \\
\midrule
\rowcolor{lightrow} DemonAttack & A fixed-shooter game where players control a cannon at the bottom of the screen defending against waves of demons descending from the top. Each wave features demons with different movement patterns, attack strategies, and point values. Some demons split into two when hit, while others drop bombs or dive-bomb the player. The player's cannon can move horizontally and fire upward. As levels progress, demons become faster and more aggressive, with more complex attack patterns. The game features distinctive sound effects and colorful graphics. \\
\bottomrule
\end{tabularx}
\end{adjustbox}
\caption{
\textbf{Detailed Descriptions of Selected Atari Games (Part 1).} This table provides comprehensive descriptions of seven classic Atari games, explaining their gameplay mechanics, objectives, and distinctive features.
}
\label{table:atari_games_description1}
\end{table}

\begin{table}[!t]
\begin{adjustbox}{center, width=\textwidth}
\small
\begin{tabularx}{\textwidth}{>{\hsize=0.27\hsize}X>{\hsize=0.73\hsize}X}
\rowcolor{header} \multicolumn{2}{c}{\textbf{Detailed Description of Atari Games (Part 2)}} \\
\midrule
\textbf{Game} & \textbf{Detailed Description} \\
\midrule
\rowcolor{lightrow} Freeway & A simple but challenging game where the player controls a chicken trying to cross a ten-lane freeway filled with traffic. The goal is to reach the other side as many times as possible within the time limit. Each lane has vehicles moving at different speeds and directions. If the chicken collides with a vehicle, it is pushed backward. Players can only move up or down, requiring precise timing to navigate through gaps in traffic. The game supports two-player competitive mode where players race to get their chickens across more times than their opponent. \\
\midrule
Frostbite & An arctic-themed game where players control Frostbite Bailey, who must build an igloo by jumping on floating ice floes in a frigid river. Each time the player lands on an ice floe, it changes color and adds a block to the igloo. Once the igloo is complete, the player must reach it to advance to the next level. Hazards include bears, geese, and crabs that patrol the ice floes. The temperature gradually drops, adding time pressure. Fish and clams appear as bonus items that can be collected for extra points. \\
\midrule
\rowcolor{lightrow} Hero & A complex adventure game where players control Roderick Hero, a miner equipped with a helicopter backpack, dynamite, and a beam weapon. The objective is to navigate through multi-screen mine shafts to rescue trapped miners. Players must blast through walls with dynamite, defeat creatures with the beam weapon, and avoid hazards like magma and falling rocks. The helicopter backpack allows limited flight but consumes power. Lanterns throughout the mine provide light in dark areas and extra power when collected. The game features complex, non-linear level designs. \\
\midrule
MontezumaRevenge & A notoriously challenging platformer set in an Aztec temple. Players control an explorer named Panama Joe navigating through multiple screens of the temple to collect treasures. The game features locked doors requiring keys, deadly traps including fire pits and rolling skulls, and enemies like snakes and spiders. Players must use ladders, ropes, and disappearing floors to navigate between rooms. The game is known for its punishing difficulty, sparse rewards, and the need for precise timing and planning. Death results in returning to the starting room, making progress particularly demanding. \\
\midrule
\rowcolor{lightrow} MsPacman & An enhanced version of Pac-Man featuring a female protagonist. Players navigate through four different maze designs consuming dots while avoiding ghosts. Power pellets allow temporary ghost consumption. Ms. Pac-Man improved upon the original with more varied mazes, ghosts with less predictable movement patterns, and bonus fruits that move around the maze rather than staying stationary. Between levels, brief cutscenes tell the love story between Ms. Pac-Man and Pac-Man. The game's less deterministic ghost behavior makes it more challenging and less susceptible to pattern-based strategies than the original. \\
\midrule
Pitfall & A groundbreaking side-scrolling platformer where players control Pitfall Harry through a jungle collecting treasures within a $20$-minute time limit. The jungle consists of $255$ interconnected screens with various hazards including rolling logs, crocodiles, scorpions, quicksand, tar pits, and fires. Players navigate by running, jumping, and swinging on vines. Underground passages allow for faster travel between areas. The game was revolutionary for its era due to its large game world and fluid character animation. Points are scored by collecting treasures, with time penalties for falling into hazards. \\
\midrule
\rowcolor{lightrow} Pong & One of the earliest and most iconic video games, simulating table tennis. Players control vertical paddles on opposite sides of the screen and must hit a ball back and forth. The ball bounces off the top and bottom edges, and points are scored when one player fails to return the ball. The ball's speed increases after several successful returns, increasing difficulty. The angle of the ball's rebound depends on which part of the paddle it hits, allowing for strategic aiming. Despite its simplicity, Pong established many foundational elements of video games and interactive entertainment. \\
\bottomrule
\end{tabularx}
\end{adjustbox}
\caption{
\textbf{Detailed Descriptions of Selected Atari Games (Part 2).} This table provides comprehensive descriptions of seven classic Atari games, explaining their gameplay mechanics, objectives, and distinctive features.
}
\label{table:atari_games_description2}
\end{table}

\begin{table}[!t]
\begin{adjustbox}{center, width=\textwidth}
\small
\begin{tabularx}{\textwidth}{>{\hsize=0.27\hsize}X>{\hsize=0.73\hsize}X}
\rowcolor{header} \multicolumn{2}{c}{\textbf{Detailed Description of Atari Games (Part 3)}} \\
\midrule
\textbf{Game} & \textbf{Detailed Description} \\
\midrule
\rowcolor{lightrow} PrivateEye & A detective adventure game where players control Private Eye Pierre Touché solving cases by navigating a scrolling city environment. The main case involves recovering stolen items from the Goldfish Diamond case. Players drive a car through the city, entering buildings, and collecting evidence while avoiding gangsters and other hazards. The game features a complex scoring system based on catching criminals, recovering stolen items, and completing the case within the time limit. Incorrect accusations result in penalties. The game's non-linear design with multiple locations to explore was innovative for its time. \\
\midrule
Qbert & A puzzle-platformer where players control Q*bert, an orange creature with a tubular nose, who hops around a pyramid of cubes. The objective is to change the color of each cube's top surface by hopping on it. Once all cubes are changed to the target color, the player advances to the next level. Enemies include Coily the snake, Wrong-Way and Ugg who travel along the sides of the pyramid, and red balls that fall from the top. Q*bert can use floating discs to escape to the top of the pyramid or to lure Coily off the edge. Later levels require changing each cube's color multiple times or in specific sequences. \\
\midrule
\rowcolor{lightrow} Riverraid & A vertically scrolling shooter where players pilot a fighter jet over a river, destroying enemy helicopters, ships, fuel depots, and bridges while avoiding collisions with the shoreline. The plane consumes fuel continuously, requiring players to fly over fuel depots to refill. The river varies in width, creating narrow passages that demand precise navigation. Destroying bridges marks progression to new sections with increased difficulty. The game was notable for its use of a pseudo-random algorithm to generate the river layout, creating a different experience each play while using minimal memory. \\
\midrule
Seaquest & An underwater shooter where players control a submarine that must rescue divers while defending against sharks, enemy submarines, and other sea creatures. The submarine can move in all directions and fire torpedoes horizontally. Players must manage their oxygen supply, surfacing periodically to replenish it and deliver rescued divers. Each surfacing with a full complement of divers awards bonus points. If oxygen depletes completely, the submarine is lost. As levels progress, enemies become more numerous and aggressive. Balancing rescue operations with defense and oxygen management creates a multi-objective challenge. \\
\midrule
\rowcolor{lightrow} Skiing & A downhill skiing simulation where players navigate through a series of gates on a continuously scrolling course. The objective is to complete the course in the shortest time possible without missing gates, which incur time penalties. Players control their skier's horizontal position as they automatically move downward. Obstacles include trees and moguls that must be avoided. The game offers two modes: the slalom, with wider gate spacing, and the more challenging giant slalom with tighter gates. Precise control and planning the optimal line through gates are essential for achieving the best times. \\
\midrule
SpaceInvaders & A fixed shooter game where players control a laser cannon moving horizontally at the bottom of the screen, defending against rows of descending aliens. The aliens move side to side, dropping bombs as they advance toward the bottom. Players must eliminate all aliens before they reach the ground. Protective bunkers provide temporary cover but degrade when hit by either player shots or alien bombs. As aliens are destroyed, the remaining invaders move faster. Strategic gameplay involves targeting specific aliens to manipulate their movement patterns and using the bunkers effectively for protection. \\
\midrule
\rowcolor{lightrow} Tennis & A sports simulation of tennis from a side view of the court. Players control tennis players who can move around their side of the court and swing rackets to hit the ball over the net. The game implements basic tennis rules including serves, volleys, and scoring ($15$-$30$-$40$-game). Ball physics include appropriate bouncing and speed changes based on the type of hit. Strategic gameplay involves positioning for shots, timing racket swings correctly, and anticipating opponent movements. The game can be played against the computer or another human player, with varying difficulty levels for the AI opponent. \\
\bottomrule
\end{tabularx}
\end{adjustbox}
\caption{
\textbf{Detailed Descriptions of Selected Atari Games (Part 3).} This table provides comprehensive descriptions of seven classic Atari games, explaining their gameplay mechanics, objectives, and distinctive features.
}
\label{table:atari_games_description3}
\end{table}

\begin{table}[!t]
\begin{adjustbox}{center, width=\textwidth}
\small
\begin{tabularx}{\textwidth}{>{\hsize=0.27\hsize}X>{\hsize=0.73\hsize}X}
\rowcolor{header} \multicolumn{2}{c}{\textbf{Detailed Description of Atari Games (Part 4)}} \\
\midrule
\textbf{Game} & \textbf{Detailed Description} \\
\midrule
\rowcolor{lightrow} Venture & An exploration game where players control Winky, an adventurer navigating through a multi-room dungeon to collect treasures. Each room contains different monsters guarding treasure, requiring specific strategies to overcome. Players view the dungeon layout from an overhead perspective but transition to a zoomed-in view when entering a room. If players take too long in a room, the invincible "Hallmonster" appears, forcing swift action. The game features four different dungeons with increasing difficulty and unique monsters in each room, from snakes and trolls to giant spiders and the Grim Reaper. \\
\midrule
VideoPinball & A digital recreation of pinball that simulates the physics and features of a traditional pinball machine. Players control left and right flippers to keep the ball in play, aiming to hit various targets to score points. The table includes bumpers, spinners, rollover targets, and bonus areas. Players can tilt the table (with limits) to influence ball direction. The game features realistic ball physics including momentum, ricochet angles, and speed changes. Special features include multiball play and bonus rounds that can be activated through specific target combinations. Scoring emphasizes both quick reflexes and strategic target selection. \\
\bottomrule
\end{tabularx}
\end{adjustbox}
\caption{
\textbf{Detailed Descriptions of Selected Atari Games (Part 4).} This table provides comprehensive descriptions of the remaining classic Atari games, explaining their gameplay mechanics, objectives, and distinctive features.
}
\label{table:atari_games_description4}
\end{table}

Together, these tables provide a thorough understanding of the task space encompassed by TextAtari, highlighting the diverse challenges that make this benchmark particularly valuable for evaluating long-horizon reasoning in language agents. 
The wide variety of game mechanics, objectives, and difficulty levels ensures that TextAtari tests a broad spectrum of reasoning capabilities essential for advanced sequential decision-making.

\section{Prompt Engineering}\label{sec:prompt}

This section provides detailed information about the prompt templates used across our TextAtari experimental framework. 
These prompts determine how information is presented to the language models and how reasoning is structured during gameplay.

\subsection{Scenario-Based Prompts}

We developed four distinct scenario-based prompts to systematically investigate how different forms of prior knowledge affect language agent performance in long-horizon game environments.

\begin{figure}[!h]
\centering
\begin{minipage}{0.95\linewidth}
\begin{promptbox}{Basic}
You are an expert-level game player. Your whole response should be in JSON format.
You are in a game. {game_description} {goal_description}

Currently, {state_description}. {action_description}
Please suggest one valid action identifier. Your Suggested Action is:
\end{promptbox}
\end{minipage}
\end{figure}

The Basic scenario provides minimal information, representing our control condition. 
The prompt includes only essential game elements: 
a brief game description, goal statement, current state description, and available actions. This lean template forces the language model to rely entirely on its intrinsic knowledge and reasoning capabilities without external guidance. 
By intentionally omitting additional context, demonstrations, or strategic hints, we establish a baseline measurement of the model's inherent game-playing abilities.

\begin{figure}[!h]
\centering
\begin{minipage}{0.95\linewidth}
\begin{promptbox}{Obscured}
You are an expert-level game player. Your whole response should be in JSON format.
You are in a game. {game_description} {goal_description}

Currently, {masked_state_description}. {action_description}
Please suggest one valid action identifier. Your Suggested Action is:
\end{promptbox}
\end{minipage}
\end{figure}

The Obscured scenario tests the model's reliance on domain-specific terminology and semantic priors. 
This prompt systematically replaces all game-specific nouns (e.g., ``ghost,'' ``paddle,'' ``asteroid'') with the generic token ``item'' while preserving structural information like coordinates, colors, and scores. 
This transformation forces the model to reason about game dynamics based purely on spatial relationships and numeric patterns rather than leveraging pre-trained associations with familiar game elements. 
The contrast between Basic and Obscured performance reveals how much models depend on semantic understanding versus structural reasoning.

\begin{figure}[!h]
\centering
\begin{minipage}{0.95\linewidth}
\begin{promptbox}{Manual Augmentation}
You are in a game. {game_description} {goal_description}

This is the game manual for this game. You need to read it carefully and understand the content and play strategies of the game:

{manual_excerpt}
\end{promptbox}
\end{minipage}
\end{figure}

The Manual Augmentation scenario supplements the basic prompt with explicit game knowledge through official game manual excerpts. 
We curate these manual snippets from authentic Atari documentation (primarily sourced from AtariAge archives), limiting them to approximately $300$ tokens to ensure consistency across games. 
These excerpts provide formal explanations of game mechanics, scoring systems, strategic considerations, and win conditions. 
This condition tests whether explicit domain knowledge injection enhances performance and to what degree different language models can operationalize written instructions into effective gameplay strategies.

\begin{figure}[!h]
\centering
\begin{minipage}{0.95\linewidth}
\begin{promptbox}{Reference-based}
You are in a game. {game_description} {goal_description}

This is the trajectory of playing this game using the RL algorithm. Please read these trajectories carefully and refer to them to make decisions during gameplay:

[{state_1} -> {action_1}; {state_2} -> {action_2}; ...]
\end{promptbox}
\end{minipage}
\end{figure}

The Reference-based scenario provides the language model with exemplar gameplay through expert demonstrations. 
For each game, we include trajectory samples from a trained Proximal Policy Optimization (PPO) agent that achieves at least average human-level performance. 
These trajectories are sampled at regular intervals (every $10$th state-action pair) and formatted as state-action sequences, giving the language model concrete examples of successful gameplay without directly encoding future knowledge. 
This condition examines how effectively language models can learn from demonstrations and adapt observed strategies to novel game states.

\subsection{Agent Reasoning Frameworks}

Beyond the knowledge variation in scenarios, we implemented three distinct reasoning frameworks that structure how language models approach decision-making.

\begin{figure}[!h]
\centering
\begin{minipage}{0.95\linewidth}
\begin{tcolorbox}[title=\textbf{Basic Agent Prompt}, colback=gray!5, colframe=gray!50, fonttitle=\bfseries]
\begin{lstlisting}[breaklines=true, basicstyle=\ttfamily\small]
{state_description}.{action_description}
Please suggest an action based on the current game state and the information you get.
You must select the appropriate action from the given action descriptions and cannot refrain from taking action or perform any prohibited actions.
Your Suggested Action is:
\end{lstlisting}
\end{tcolorbox}
\end{minipage}
\end{figure}

The Basic Agent employs the simplest decision-making process, requesting a direct action selection without intermediate reasoning steps. 
The prompt instructs the model to suggest a valid action based on the current game state, emphasizing the need to select from available actions without abstention. 
This framework tests the model's ability to make intuitive decisions without explicit reasoning prompts, relying on the model's internal processing to map observations directly to actions.

\begin{figure}[!h]
\centering
\begin{minipage}{0.95\linewidth}
\begin{tcolorbox}[title=\textbf{Chain-of-Thought Agent Prompt}, colback=gray!5, colframe=gray!50, fonttitle=\bfseries]
\begin{lstlisting}[breaklines=true, basicstyle=\ttfamily\small]
Currently, {state_description}. {action_description}
Now select your action. You should first take a deep breath. Then you should think step by step about the action selection and lay out your thought process explicitly. After that you should decide an action based on the thought. For the whole response, you should use JSON format with two keys "thought process" and "action".
\end{lstlisting}
\end{tcolorbox}
\end{minipage}
\end{figure}

The Chain-of-Thought (CoT) Agent introduces explicit reasoning into the decision process by requiring the model to articulate its thought process before selecting an action. 
The prompt instructs the model to ``take a deep breath'' and ``think step by step,'' encouraging deliberate consideration of the current state, available actions, and potential outcomes. 
The response must follow a structured JSON format with separate fields for the thought process and final action selection. 
This framework tests whether explicit reasoning improves decision quality in long-horizon gaming environments.

\begin{figure}[!h]
\centering
\begin{minipage}{0.95\linewidth}
\begin{tcolorbox}[title=\textbf{Reflection Agent Prompt}, colback=gray!5, colframe=gray!50, fonttitle=\bfseries]
\begin{lstlisting}[breaklines=true, basicstyle=\ttfamily\small]
You are an analytic and game coach. You need to analyse the game and summarize the current strategy step by step.
You will be given the history of a past experience in which you were placed in an environment and given a task to complete. You were unsuccessful in completing the task.
Do not summarize your environment, but rather think about the strategy and path you took to attempt to complete the task. Think step by step what mistakes you made leading the failure.
Then devise a concise, new plan of action that accounts for your mistake with reference to specific actions that you should have taken.
For example, if you tried A and B but forgot C, then you should reason that forgetting C is the key mistake.
After that, devise a plan to achieve C with environment-specific actions.
Remind yourself of the plan you will take in the next trial and give your plan after "Plan".
Respond in JSON with four keys: "Strategy", "Knowledge", "Reflexion", and "New Plan".
\end{lstlisting}
\end{tcolorbox}
\end{minipage}
\end{figure}

The Reflection Agent extends the CoT framework by incorporating episodic self-critique and strategic adaptation. 
After each game episode concludes, this agent prompts the model to analyze its performance, identify mistakes, extract lessons, and formulate an improved strategy for subsequent attempts. 
The reflection must follow a structured JSON format covering strategy analysis, knowledge updates, specific reflection on failures, and a concrete new plan. 
These reflections are preserved and presented to the model at the beginning of subsequent episodes, creating a feedback loop that enables iterative improvement without parameter updates. 
This framework tests whether meta-cognitive processes enhance performance over extended interactions.

The combination of these scenario-based knowledge variations and reasoning frameworks creates a comprehensive experimental matrix for evaluating language models' capabilities in long-horizon decision-making tasks. 
By systematically controlling these variables while maintaining consistent evaluation protocols, we can isolate the specific contributions of different knowledge sources and reasoning strategies to overall performance.

\section{Missing Results}\label{sec:missing-results}

In this section, we provide a comprehensive set of additional results and visualizations that extend the analysis presented in the main paper. 
These figures offer a detailed performance comparison across the three dimensions of our experimental framework: 
model architectures (Qwen2.5-7B, Llama3.1-8B, and Gemma-7B), scenario types (Basic, Obscured, Game Manual, and Reference-based), and agent frameworks (Naive, Chain-of-Thought, Reflexion\_last, and Reflexion\_max).

\noindent\textbf{Nomenclature Note.}
For clarity, we should note that throughout this appendix, the visualization labels use ``RL Trajectory'' to refer to what we call the ``Reference-based'' scenario in the main text. 
This naming discrepancy reflects implementation details, as the Reference-based scenario was implemented using trajectories generated by reinforcement learning (PPO) agents. 
The underlying experimental condition remains consistent with the Reference-based scenario described in the methodology section.

\subsection{Experimental Framework and Visualization Structure}

Our experimental framework explores three critical dimensions. 
First, we evaluate three state-of-the-art open-source small-scale LLMs - Qwen2.5-7B, Llama3.1-8B, and Gemma-7B. 
These models represent different architectural approaches, training methodologies, and capabilities, allowing us to examine how fundamental model design affects performance on long-horizon sequential decision-making tasks.

Second, we investigate four distinct knowledge conditions. 
The Basic scenario provides only the essential game description and current observation. 
The Obscured scenario replaces domain-specific nouns with neutral tokens to test reliance on lexical priors. 
The Game Manual scenario supplements the Basic scenario with concise game manual excerpts. The Reference-based scenario (labeled as ``RL Trajectory'' in figures) primes the agent with expert demonstrations generated by reinforcement learning algorithms.

Third, we implement four prompting strategies. 
The Naive agent employs a zero-shot approach with minimal prompting. 
The Chain-of-Thought (CoT) agent encourages step-by-step reasoning before making decisions. The Reflexion\_last agent incorporates the most recent episode reflection to guide current gameplay. 
The Reflexion\_max agent utilizes the best-performing reflection from previous episodes to optimize decision-making.

Each figure in this appendix presents a systematic comparison while holding two dimensions fixed and varying the third. 
The visualizations use a consistent format wherein bar charts represent the relative performance differences (percentage change) across all 23 Atari environments, while vertical lines indicate the normalized average performance scores for each configuration. 
Each row contains multiple pairwise comparisons, allowing for direct assessment of performance trends across different experimental conditions.

The figures are organized into three major categories. 
The first 12 figures examine how different knowledge conditions affect performance while keeping the model architecture and agent framework constant. 
The second 12 figures explore how different prompting strategies affect performance while keeping the model architecture and scenario type constant. 
The left figures investigate how different model architectures perform while keeping the scenario type and agent framework constant.

\begin{figure}[htb!]
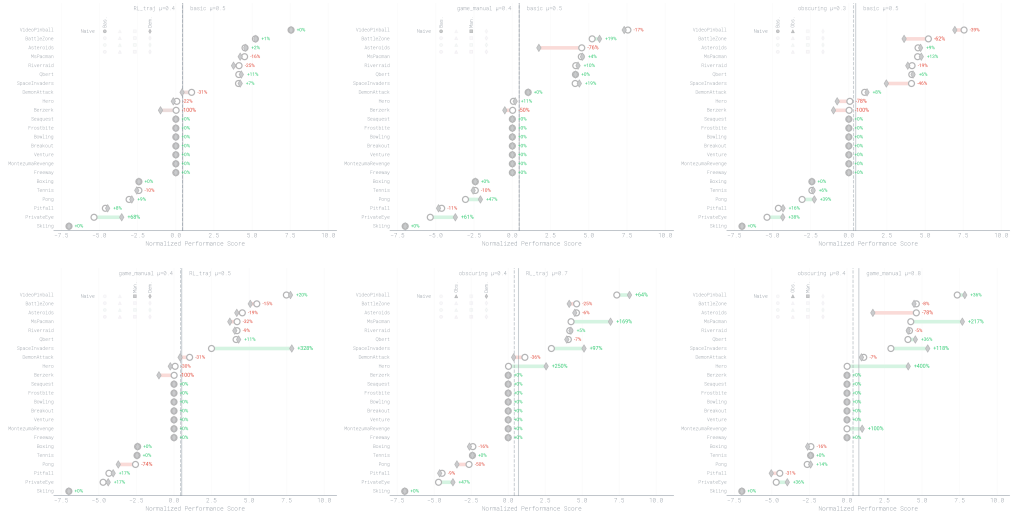

    \centering
    \includegraphics[width=0.32\textwidth]{figs-appendix-arxiv/qwen7b\_naive\_basic\_RL\_traj.png}%
    \includegraphics[width=0.32\textwidth]{figs-appendix-arxiv/qwen7b\_naive\_basic\_game\_manual.png}%
    \includegraphics[width=0.32\textwidth]{figs-appendix-arxiv/qwen7b\_naive\_basic\_obscuring.png}%
    \\[1ex]
    \includegraphics[width=0.32\textwidth]{figs-appendix-arxiv/qwen7b\_naive\_game\_manual\_RL\_traj.png}%
    \includegraphics[width=0.32\textwidth]{figs-appendix-arxiv/qwen7b\_naive\_obscuring\_RL\_traj.png}%
    \includegraphics[width=0.32\textwidth]{figs-appendix-arxiv/qwen7b\_naive\_obscuring\_game\_manual.png}%
    \\[1ex]
    \caption{Performance comparison of Qwen2.5-7B with Naive agent across different scenarios. Each plot shows relative performance differences (bars) and normalized average scores (vertical lines) across all 23 Atari environments. Top row: Comparison between (left) Basic vs. RL Trajectory, (middle) Basic vs. Game Manual, and (right) Basic vs. Obscured scenarios. Bottom row: Comparison between (left) Game Manual vs. RL Trajectory, (middle) Obscured vs. RL Trajectory, and (right) Obscured vs. Game Manual scenarios.}
\end{figure}

\begin{figure}[htb!]
    \centering
    \includegraphics[width=0.32\textwidth]{figs-appendix-arxiv/qwen7b\_cot\_basic\_RL\_traj.png}%
    \includegraphics[width=0.32\textwidth]{figs-appendix-arxiv/qwen7b\_cot\_basic\_game\_manual.png}%
    \includegraphics[width=0.32\textwidth]{figs-appendix-arxiv/qwen7b\_cot\_basic\_obscuring.png}%
    \\[1ex]
    \includegraphics[width=0.32\textwidth]{figs-appendix-arxiv/qwen7b\_cot\_game\_manual\_RL\_traj.png}%
    \includegraphics[width=0.32\textwidth]{figs-appendix-arxiv/qwen7b\_cot\_obscuring\_RL\_traj.png}%
    \includegraphics[width=0.32\textwidth]{figs-appendix-arxiv/qwen7b\_cot\_obscuring\_game\_manual.png}%
    \\[1ex]
    \caption{Performance comparison of Qwen2.5-7B with Chain-of-Thought (CoT) agent across different scenarios. Each plot shows relative performance differences (bars) and normalized average scores (vertical lines) across all 23 Atari environments. Top row: Comparison between (left) Basic vs. RL Trajectory, (middle) Basic vs. Game Manual, and (right) Basic vs. Obscured scenarios. Bottom row: Comparison between (left) Game Manual vs. RL Trajectory, (middle) Obscured vs. RL Trajectory, and (right) Obscured vs. Game Manual scenarios.}
\end{figure}

\begin{figure}[htb!]
    \centering
    \includegraphics[width=0.32\textwidth]{figs-appendix-arxiv/qwen7b\_reflexion\_last\_basic\_RL\_traj.png}%
    \includegraphics[width=0.32\textwidth]{figs-appendix-arxiv/qwen7b\_reflexion\_last\_basic\_game\_manual.png}%
    \includegraphics[width=0.32\textwidth]{figs-appendix-arxiv/qwen7b\_reflexion\_last\_basic\_obscuring.png}%
    \\[1ex]
    \includegraphics[width=0.32\textwidth]{figs-appendix-arxiv/qwen7b\_reflexion\_last\_game\_manual\_RL\_traj.png}%
    \includegraphics[width=0.32\textwidth]{figs-appendix-arxiv/qwen7b\_reflexion\_last\_obscuring\_RL\_traj.png}%
    \includegraphics[width=0.32\textwidth]{figs-appendix-arxiv/qwen7b\_reflexion\_last\_obscuring\_game\_manual.png}%
    \\[1ex]
    \caption{Performance comparison of Qwen2.5-7B with Reflexion\_last agent (using the most recent reflection) across different scenarios. Each plot shows relative performance differences (bars) and normalized average scores (vertical lines) across all 23 Atari environments. Top row: Comparison between (left) Basic vs. RL Trajectory, (middle) Basic vs. Game Manual, and (right) Basic vs. Obscured scenarios. Bottom row: Comparison between (left) Game Manual vs. RL Trajectory, (middle) Obscured vs. RL Trajectory, and (right) Obscured vs. Game Manual scenarios.}
\end{figure}

\begin{figure}[htb!]
    \centering
    \includegraphics[width=0.32\textwidth]{figs-appendix-arxiv/qwen7b\_reflexion\_max\_basic\_RL\_traj.png}%
    \includegraphics[width=0.32\textwidth]{figs-appendix-arxiv/qwen7b\_reflexion\_max\_basic\_game\_manual.png}%
    \includegraphics[width=0.32\textwidth]{figs-appendix-arxiv/qwen7b\_reflexion\_max\_basic\_obscuring.png}%
    \\[1ex]
    \includegraphics[width=0.32\textwidth]{figs-appendix-arxiv/qwen7b\_reflexion\_max\_game\_manual\_RL\_traj.png}%
    \includegraphics[width=0.32\textwidth]{figs-appendix-arxiv/qwen7b\_reflexion\_max\_obscuring\_RL\_traj.png}%
    \includegraphics[width=0.32\textwidth]{figs-appendix-arxiv/qwen7b\_reflexion\_max\_obscuring\_game\_manual.png}%
    \\[1ex]
    \caption{Performance comparison of Qwen2.5-7B with Reflexion\_max agent (using the best-performing reflection) across different scenarios. Each plot shows relative performance differences (bars) and normalized average scores (vertical lines) across all 23 Atari environments. Top row: Comparison between (left) Basic vs. RL Trajectory, (middle) Basic vs. Game Manual, and (right) Basic vs. Obscured scenarios. Bottom row: Comparison between (left) Game Manual vs. RL Trajectory, (middle) Obscured vs. RL Trajectory, and (right) Obscured vs. Game Manual scenarios.}
\end{figure}

\begin{figure}[htb!]
    \centering
    \includegraphics[width=0.32\textwidth]{figs-appendix-arxiv/llama\_naive\_basic\_RL\_traj.png}%
    \includegraphics[width=0.32\textwidth]{figs-appendix-arxiv/llama\_naive\_basic\_game\_manual.png}%
    \includegraphics[width=0.32\textwidth]{figs-appendix-arxiv/llama\_naive\_basic\_obscuring.png}%
    \\[1ex]
    \includegraphics[width=0.32\textwidth]{figs-appendix-arxiv/llama\_naive\_game\_manual\_RL\_traj.png}%
    \includegraphics[width=0.32\textwidth]{figs-appendix-arxiv/llama\_naive\_obscuring\_RL\_traj.png}%
    \includegraphics[width=0.32\textwidth]{figs-appendix-arxiv/llama\_naive\_obscuring\_game\_manual.png}%
    \\[1ex]
    \caption{Performance comparison of Llama3.1-8B with Naive agent across different scenarios. Each plot shows relative performance differences (bars) and normalized average scores (vertical lines) across all 23 Atari environments. Top row: Comparison between (left) Basic vs. RL Trajectory, (middle) Basic vs. Game Manual, and (right) Basic vs. Obscured scenarios. Bottom row: Comparison between (left) Game Manual vs. RL Trajectory, (middle) Obscured vs. RL Trajectory, and (right) Obscured vs. Game Manual scenarios.}
\end{figure}

\begin{figure}[htb!]
    \centering
    \includegraphics[width=0.32\textwidth]{figs-appendix-arxiv/llama\_cot\_basic\_RL\_traj.png}%
    \includegraphics[width=0.32\textwidth]{figs-appendix-arxiv/llama\_cot\_basic\_game\_manual.png}%
    \includegraphics[width=0.32\textwidth]{figs-appendix-arxiv/llama\_cot\_basic\_obscuring.png}%
    \\[1ex]
    \includegraphics[width=0.32\textwidth]{figs-appendix-arxiv/llama\_cot\_game\_manual\_RL\_traj.png}%
    \includegraphics[width=0.32\textwidth]{figs-appendix-arxiv/llama\_cot\_obscuring\_RL\_traj.png}%
    \includegraphics[width=0.32\textwidth]{figs-appendix-arxiv/llama\_cot\_obscuring\_game\_manual.png}%
    \\[1ex]
    \caption{Performance comparison of Llama3.1-8B with Chain-of-Thought (CoT) agent across different scenarios. Each plot shows relative performance differences (bars) and normalized average scores (vertical lines) across all 23 Atari environments. Top row: Comparison between (left) Basic vs. RL Trajectory, (middle) Basic vs. Game Manual, and (right) Basic vs. Obscured scenarios. Bottom row: Comparison between (left) Game Manual vs. RL Trajectory, (middle) Obscured vs. RL Trajectory, and (right) Obscured vs. Game Manual scenarios.}
\end{figure}

\begin{figure}[htb!]
    \centering
    \includegraphics[width=0.32\textwidth]{figs-appendix-arxiv/llama\_reflexion\_last\_basic\_RL\_traj.png}%
    \includegraphics[width=0.32\textwidth]{figs-appendix-arxiv/llama\_reflexion\_last\_basic\_game\_manual.png}%
    \includegraphics[width=0.32\textwidth]{figs-appendix-arxiv/llama\_reflexion\_last\_basic\_obscuring.png}%
    \\[1ex]
    \includegraphics[width=0.32\textwidth]{figs-appendix-arxiv/llama\_reflexion\_last\_game\_manual\_RL\_traj.png}%
    \includegraphics[width=0.32\textwidth]{figs-appendix-arxiv/llama\_reflexion\_last\_obscuring\_RL\_traj.png}%
    \includegraphics[width=0.32\textwidth]{figs-appendix-arxiv/llama\_reflexion\_last\_obscuring\_game\_manual.png}%
    \\[1ex]
    \caption{Performance comparison of Llama3.1-8B with Reflexion\_last agent (using the most recent reflection) across different scenarios. Each plot shows relative performance differences (bars) and normalized average scores (vertical lines) across all 23 Atari environments. Top row: Comparison between (left) Basic vs. RL Trajectory, (middle) Basic vs. Game Manual, and (right) Basic vs. Obscured scenarios. Bottom row: Comparison between (left) Game Manual vs. RL Trajectory, (middle) Obscured vs. RL Trajectory, and (right) Obscured vs. Game Manual scenarios.}
\end{figure}

\begin{figure}[htb!]
    \centering
    \includegraphics[width=0.32\textwidth]{figs-appendix-arxiv/llama\_reflexion\_max\_basic\_RL\_traj.png}%
    \includegraphics[width=0.32\textwidth]{figs-appendix-arxiv/llama\_reflexion\_max\_basic\_game\_manual.png}%
    \includegraphics[width=0.32\textwidth]{figs-appendix-arxiv/llama\_reflexion\_max\_basic\_obscuring.png}%
    \\[1ex]
    \includegraphics[width=0.32\textwidth]{figs-appendix-arxiv/llama\_reflexion\_max\_game\_manual\_RL\_traj.png}%
    \includegraphics[width=0.32\textwidth]{figs-appendix-arxiv/llama\_reflexion\_max\_obscuring\_RL\_traj.png}%
    \includegraphics[width=0.32\textwidth]{figs-appendix-arxiv/llama\_reflexion\_max\_obscuring\_game\_manual.png}%
    \\[1ex]
    \caption{Performance comparison of Llama3.1-8B with Reflexion\_max agent (using the best-performing reflection) across different scenarios. Each plot shows relative performance differences (bars) and normalized average scores (vertical lines) across all 23 Atari environments. Top row: Comparison between (left) Basic vs. RL Trajectory, (middle) Basic vs. Game Manual, and (right) Basic vs. Obscured scenarios. Bottom row: Comparison between (left) Game Manual vs. RL Trajectory, (middle) Obscured vs. RL Trajectory, and (right) Obscured vs. Game Manual scenarios.}
\end{figure}

\begin{figure}[htb!]
    \centering
    \includegraphics[width=0.32\textwidth]{figs-appendix-arxiv/gemma\_naive\_basic\_RL\_traj.png}%
    \includegraphics[width=0.32\textwidth]{figs-appendix-arxiv/gemma\_naive\_basic\_game\_manual.png}%
    \includegraphics[width=0.32\textwidth]{figs-appendix-arxiv/gemma\_naive\_basic\_obscuring.png}%
    \\[1ex]
    \includegraphics[width=0.32\textwidth]{figs-appendix-arxiv/gemma\_naive\_game\_manual\_RL\_traj.png}%
    \includegraphics[width=0.32\textwidth]{figs-appendix-arxiv/gemma\_naive\_obscuring\_RL\_traj.png}%
    \includegraphics[width=0.32\textwidth]{figs-appendix-arxiv/gemma\_naive\_obscuring\_game\_manual.png}%
    \\[1ex]
    \caption{Performance comparison of Gemma-7B with Naive agent across different scenarios. Each plot shows relative performance differences (bars) and normalized average scores (vertical lines) across all 23 Atari environments. Top row: Comparison between (left) Basic vs. RL Trajectory, (middle) Basic vs. Game Manual, and (right) Basic vs. Obscured scenarios. Bottom row: Comparison between (left) Game Manual vs. RL Trajectory, (middle) Obscured vs. RL Trajectory, and (right) Obscured vs. Game Manual scenarios.}
\end{figure}

\begin{figure}[htb!]
    \centering
    \includegraphics[width=0.32\textwidth]{figs-appendix-arxiv/gemma\_cot\_basic\_RL\_traj.png}%
    \includegraphics[width=0.32\textwidth]{figs-appendix-arxiv/gemma\_cot\_basic\_game\_manual.png}%
    \includegraphics[width=0.32\textwidth]{figs-appendix-arxiv/gemma\_cot\_basic\_obscuring.png}%
    \\[1ex]
    \includegraphics[width=0.32\textwidth]{figs-appendix-arxiv/gemma\_cot\_game\_manual\_RL\_traj.png}%
    \includegraphics[width=0.32\textwidth]{figs-appendix-arxiv/gemma\_cot\_obscuring\_RL\_traj.png}%
    \includegraphics[width=0.32\textwidth]{figs-appendix-arxiv/gemma\_cot\_obscuring\_game\_manual.png}%
    \\[1ex]
    \caption{Performance comparison of Gemma-7B with Chain-of-Thought (CoT) agent across different scenarios. Each plot shows relative performance differences (bars) and normalized average scores (vertical lines) across all 23 Atari environments. Top row: Comparison between (left) Basic vs. RL Trajectory, (middle) Basic vs. Game Manual, and (right) Basic vs. Obscured scenarios. Bottom row: Comparison between (left) Game Manual vs. RL Trajectory, (middle) Obscured vs. RL Trajectory, and (right) Obscured vs. Game Manual scenarios.}
\end{figure}

\begin{figure}[htb!]
    \centering
    \includegraphics[width=0.32\textwidth]{figs-appendix-arxiv/gemma\_reflexion\_last\_basic\_RL\_traj.png}%
    \includegraphics[width=0.32\textwidth]{figs-appendix-arxiv/gemma\_reflexion\_last\_basic\_game\_manual.png}%
    \includegraphics[width=0.32\textwidth]{figs-appendix-arxiv/gemma\_reflexion\_last\_basic\_obscuring.png}%
    \\[1ex]
    \includegraphics[width=0.32\textwidth]{figs-appendix-arxiv/gemma\_reflexion\_last\_game\_manual\_RL\_traj.png}%
    \includegraphics[width=0.32\textwidth]{figs-appendix-arxiv/gemma\_reflexion\_last\_obscuring\_RL\_traj.png}%
    \includegraphics[width=0.32\textwidth]{figs-appendix-arxiv/gemma\_reflexion\_last\_obscuring\_game\_manual.png}%
    \\[1ex]
    \caption{Performance comparison of Gemma-7B with Reflexion\_last agent (using the most recent reflection) across different scenarios. Each plot shows relative performance differences (bars) and normalized average scores (vertical lines) across all 23 Atari environments. Top row: Comparison between (left) Basic vs. RL Trajectory, (middle) Basic vs. Game Manual, and (right) Basic vs. Obscured scenarios. Bottom row: Comparison between (left) Game Manual vs. RL Trajectory, (middle) Obscured vs. RL Trajectory, and (right) Obscured vs. Game Manual scenarios.}
\end{figure}

\begin{figure}[htb!]
    \centering
    \includegraphics[width=0.32\textwidth]{figs-appendix-arxiv/gemma\_reflexion\_max\_basic\_RL\_traj.png}%
    \includegraphics[width=0.32\textwidth]{figs-appendix-arxiv/gemma\_reflexion\_max\_basic\_game\_manual.png}%
    \includegraphics[width=0.32\textwidth]{figs-appendix-arxiv/gemma\_reflexion\_max\_basic\_obscuring.png}%
    \\[1ex]
    \includegraphics[width=0.32\textwidth]{figs-appendix-arxiv/gemma\_reflexion\_max\_game\_manual\_RL\_traj.png}%
    \includegraphics[width=0.32\textwidth]{figs-appendix-arxiv/gemma\_reflexion\_max\_obscuring\_RL\_traj.png}%
    \includegraphics[width=0.32\textwidth]{figs-appendix-arxiv/gemma\_reflexion\_max\_obscuring\_game\_manual.png}%
    \\[1ex]
    \caption{Performance comparison of Gemma-7B with Reflexion\_max agent (using the best-performing reflection) across different scenarios. Each plot shows relative performance differences (bars) and normalized average scores (vertical lines) across all 23 Atari environments. Top row: Comparison between (left) Basic vs. RL Trajectory, (middle) Basic vs. Game Manual, and (right) Basic vs. Obscured scenarios. Bottom row: Comparison between (left) Game Manual vs. RL Trajectory, (middle) Obscured vs. RL Trajectory, and (right) Obscured vs. Game Manual scenarios.}
\end{figure}

\begin{figure}[htb!]
    \centering
    \includegraphics[width=0.32\textwidth]{figs-appendix-arxiv/qwen7b\_basic\_cot\_reflexion\_last.png}%
    \includegraphics[width=0.32\textwidth]{figs-appendix-arxiv/qwen7b\_basic\_cot\_reflexion\_max.png}%
    \includegraphics[width=0.32\textwidth]{figs-appendix-arxiv/qwen7b\_basic\_naive\_cot.png}%
    \\[1ex]
    \includegraphics[width=0.32\textwidth]{figs-appendix-arxiv/qwen7b\_basic\_naive\_reflexion\_last.png}%
    \includegraphics[width=0.32\textwidth]{figs-appendix-arxiv/qwen7b\_basic\_naive\_reflexion\_max.png}%
    \includegraphics[width=0.32\textwidth]{figs-appendix-arxiv/qwen7b\_basic\_reflexion\_last\_reflexion\_max.png}%
    \\[1ex]
    \caption{Performance comparison of Qwen2.5-7B in the Basic scenario across different agent types. Each plot shows relative performance differences (bars) and normalized average scores (vertical lines) across all 23 Atari environments. Top row: Comparison between (left) CoT vs. Reflexion\_last, (middle) CoT vs. Reflexion\_max, and (right) Naive vs. CoT agents. Bottom row: Comparison between (left) Naive vs. Reflexion\_last, (middle) Naive vs. Reflexion\_max, and (right) Reflexion\_last vs. Reflexion\_max agents.}
\end{figure}

\begin{figure}[htb!]
    \centering
    \includegraphics[width=0.32\textwidth]{figs-appendix-arxiv/qwen7b\_obscuring\_cot\_reflexion\_last.png}%
    \includegraphics[width=0.32\textwidth]{figs-appendix-arxiv/qwen7b\_obscuring\_cot\_reflexion\_max.png}%
    \includegraphics[width=0.32\textwidth]{figs-appendix-arxiv/qwen7b\_obscuring\_naive\_cot.png}%
    \\[1ex]
    \includegraphics[width=0.32\textwidth]{figs-appendix-arxiv/qwen7b\_obscuring\_naive\_reflexion\_last.png}%
    \includegraphics[width=0.32\textwidth]{figs-appendix-arxiv/qwen7b\_obscuring\_naive\_reflexion\_max.png}%
    \includegraphics[width=0.32\textwidth]{figs-appendix-arxiv/qwen7b\_obscuring\_reflexion\_last\_reflexion\_max.png}%
    \\[1ex]
    \caption{Performance comparison of Qwen2.5-7B in the Obscured scenario across different agent types. Each plot shows relative performance differences (bars) and normalized average scores (vertical lines) across all 23 Atari environments. Top row: Comparison between (left) CoT vs. Reflexion\_last, (middle) CoT vs. Reflexion\_max, and (right) Naive vs. CoT agents. Bottom row: Comparison between (left) Naive vs. Reflexion\_last, (middle) Naive vs. Reflexion\_max, and (right) Reflexion\_last vs. Reflexion\_max agents.}
\end{figure}

\begin{figure}[htb!]
    \centering
    \includegraphics[width=0.32\textwidth]{figs-appendix-arxiv/qwen7b\_game\_manual\_cot\_reflexion\_last.png}%
    \includegraphics[width=0.32\textwidth]{figs-appendix-arxiv/qwen7b\_game\_manual\_cot\_reflexion\_max.png}%
    \includegraphics[width=0.32\textwidth]{figs-appendix-arxiv/qwen7b\_game\_manual\_naive\_cot.png}%
    \\[1ex]
    \includegraphics[width=0.32\textwidth]{figs-appendix-arxiv/qwen7b\_game\_manual\_naive\_reflexion\_last.png}%
    \includegraphics[width=0.32\textwidth]{figs-appendix-arxiv/qwen7b\_game\_manual\_naive\_reflexion\_max.png}%
    \includegraphics[width=0.32\textwidth]{figs-appendix-arxiv/qwen7b\_game\_manual\_reflexion\_last\_reflexion\_max.png}%
    \\[1ex]
    \caption{Performance comparison of Qwen2.5-7B in the Game Manual scenario across different agent types. Each plot shows relative performance differences (bars) and normalized average scores (vertical lines) across all 23 Atari environments. Top row: Comparison between (left) CoT vs. Reflexion\_last, (middle) CoT vs. Reflexion\_max, and (right) Naive vs. CoT agents. Bottom row: Comparison between (left) Naive vs. Reflexion\_last, (middle) Naive vs. Reflexion\_max, and (right) Reflexion\_last vs. Reflexion\_max agents.}
\end{figure}

\begin{figure}[htb!]
    \centering
    \includegraphics[width=0.32\textwidth]{figs-appendix-arxiv/qwen7b\_RL\_traj\_cot\_reflexion\_last.png}%
    \includegraphics[width=0.32\textwidth]{figs-appendix-arxiv/qwen7b\_RL\_traj\_cot\_reflexion\_max.png}%
    \includegraphics[width=0.32\textwidth]{figs-appendix-arxiv/qwen7b\_RL\_traj\_naive\_cot.png}%
    \\[1ex]
    \includegraphics[width=0.32\textwidth]{figs-appendix-arxiv/qwen7b\_RL\_traj\_naive\_reflexion\_last.png}%
    \includegraphics[width=0.32\textwidth]{figs-appendix-arxiv/qwen7b\_RL\_traj\_naive\_reflexion\_max.png}%
    \includegraphics[width=0.32\textwidth]{figs-appendix-arxiv/qwen7b\_RL\_traj\_reflexion\_last\_reflexion\_max.png}%
    \\[1ex]
    \caption{Performance comparison of Qwen2.5-7B in the RL Trajectory scenario across different agent types. Each plot shows relative performance differences (bars) and normalized average scores (vertical lines) across all 23 Atari environments. Top row: Comparison between (left) CoT vs. Reflexion\_last, (middle) CoT vs. Reflexion\_max, and (right) Naive vs. CoT agents. Bottom row: Comparison between (left) Naive vs. Reflexion\_last, (middle) Naive vs. Reflexion\_max, and (right) Reflexion\_last vs. Reflexion\_max agents.}
\end{figure}

\begin{figure}[htb!]
    \centering
    \includegraphics[width=0.32\textwidth]{figs-appendix-arxiv/llama\_basic\_cot\_reflexion\_last.png}%
    \includegraphics[width=0.32\textwidth]{figs-appendix-arxiv/llama\_basic\_cot\_reflexion\_max.png}%
    \includegraphics[width=0.32\textwidth]{figs-appendix-arxiv/llama\_basic\_naive\_cot.png}%
    \\[1ex]
    \includegraphics[width=0.32\textwidth]{figs-appendix-arxiv/llama\_basic\_naive\_reflexion\_last.png}%
    \includegraphics[width=0.32\textwidth]{figs-appendix-arxiv/llama\_basic\_naive\_reflexion\_max.png}%
    \includegraphics[width=0.32\textwidth]{figs-appendix-arxiv/llama\_basic\_reflexion\_last\_reflexion\_max.png}%
    \\[1ex]
    \caption{Performance comparison of Llama3.1-8B in the Basic scenario across different agent types. Each plot shows relative performance differences (bars) and normalized average scores (vertical lines) across all 23 Atari environments. Top row: Comparison between (left) CoT vs. Reflexion\_last, (middle) CoT vs. Reflexion\_max, and (right) Naive vs. CoT agents. Bottom row: Comparison between (left) Naive vs. Reflexion\_last, (middle) Naive vs. Reflexion\_max, and (right) Reflexion\_last vs. Reflexion\_max agents.}
\end{figure}

\begin{figure}[htb!]
    \centering
    \includegraphics[width=0.32\textwidth]{figs-appendix-arxiv/llama\_obscuring\_cot\_reflexion\_last.png}%
    \includegraphics[width=0.32\textwidth]{figs-appendix-arxiv/llama\_obscuring\_cot\_reflexion\_max.png}%
    \includegraphics[width=0.32\textwidth]{figs-appendix-arxiv/llama\_obscuring\_naive\_cot.png}%
    \\[1ex]
    \includegraphics[width=0.32\textwidth]{figs-appendix-arxiv/llama\_obscuring\_naive\_reflexion\_last.png}%
    \includegraphics[width=0.32\textwidth]{figs-appendix-arxiv/llama\_obscuring\_naive\_reflexion\_max.png}%
    \includegraphics[width=0.32\textwidth]{figs-appendix-arxiv/llama\_obscuring\_reflexion\_last\_reflexion\_max.png}%
    \\[1ex]
    \caption{Performance comparison of Llama3.1-8B in the Obscured scenario across different agent types. Each plot shows relative performance differences (bars) and normalized average scores (vertical lines) across all 23 Atari environments. Top row: Comparison between (left) CoT vs. Reflexion\_last, (middle) CoT vs. Reflexion\_max, and (right) Naive vs. CoT agents. Bottom row: Comparison between (left) Naive vs. Reflexion\_last, (middle) Naive vs. Reflexion\_max, and (right) Reflexion\_last vs. Reflexion\_max agents.}
\end{figure}

\begin{figure}[htb!]
    \centering
    \includegraphics[width=0.32\textwidth]{figs-appendix-arxiv/llama\_game\_manual\_cot\_reflexion\_last.png}%
    \includegraphics[width=0.32\textwidth]{figs-appendix-arxiv/llama\_game\_manual\_cot\_reflexion\_max.png}%
    \includegraphics[width=0.32\textwidth]{figs-appendix-arxiv/llama\_game\_manual\_naive\_cot.png}%
    \\[1ex]
    \includegraphics[width=0.32\textwidth]{figs-appendix-arxiv/llama\_game\_manual\_naive\_reflexion\_last.png}%
    \includegraphics[width=0.32\textwidth]{figs-appendix-arxiv/llama\_game\_manual\_naive\_reflexion\_max.png}%
    \includegraphics[width=0.32\textwidth]{figs-appendix-arxiv/llama\_game\_manual\_reflexion\_last\_reflexion\_max.png}%
    \\[1ex]
    \caption{Performance comparison of Llama3.1-8B in the Game Manual scenario across different agent types. Each plot shows relative performance differences (bars) and normalized average scores (vertical lines) across all 23 Atari environments. Top row: Comparison between (left) CoT vs. Reflexion\_last, (middle) CoT vs. Reflexion\_max, and (right) Naive vs. CoT agents. Bottom row: Comparison between (left) Naive vs. Reflexion\_last, (middle) Naive vs. Reflexion\_max, and (right) Reflexion\_last vs. Reflexion\_max agents.}
\end{figure}

\begin{figure}[htb!]
    \centering
    \includegraphics[width=0.32\textwidth]{figs-appendix-arxiv/llama\_RL\_traj\_cot\_reflexion\_last.png}%
    \includegraphics[width=0.32\textwidth]{figs-appendix-arxiv/llama\_RL\_traj\_cot\_reflexion\_max.png}%
    \includegraphics[width=0.32\textwidth]{figs-appendix-arxiv/llama\_RL\_traj\_naive\_cot.png}%
    \\[1ex]
    \includegraphics[width=0.32\textwidth]{figs-appendix-arxiv/llama\_RL\_traj\_naive\_reflexion\_last.png}%
    \includegraphics[width=0.32\textwidth]{figs-appendix-arxiv/llama\_RL\_traj\_naive\_reflexion\_max.png}%
    \includegraphics[width=0.32\textwidth]{figs-appendix-arxiv/llama\_RL\_traj\_reflexion\_last\_reflexion\_max.png}%
    \\[1ex]
    \caption{Performance comparison of Llama3.1-8B in the RL Trajectory scenario across different agent types. Each plot shows relative performance differences (bars) and normalized average scores (vertical lines) across all 23 Atari environments. Top row: Comparison between (left) CoT vs. Reflexion\_last, (middle) CoT vs. Reflexion\_max, and (right) Naive vs. CoT agents. Bottom row: Comparison between (left) Naive vs. Reflexion\_last, (middle) Naive vs. Reflexion\_max, and (right) Reflexion\_last vs. Reflexion\_max agents.}
\end{figure}

\begin{figure}[htb!]
    \centering
    \includegraphics[width=0.32\textwidth]{figs-appendix-arxiv/gemma\_basic\_cot\_reflexion\_last.png}%
    \includegraphics[width=0.32\textwidth]{figs-appendix-arxiv/gemma\_basic\_cot\_reflexion\_max.png}%
    \includegraphics[width=0.32\textwidth]{figs-appendix-arxiv/gemma\_basic\_naive\_cot.png}%
    \\[1ex]
    \includegraphics[width=0.32\textwidth]{figs-appendix-arxiv/gemma\_basic\_naive\_reflexion\_last.png}%
    \includegraphics[width=0.32\textwidth]{figs-appendix-arxiv/gemma\_basic\_naive\_reflexion\_max.png}%
    \includegraphics[width=0.32\textwidth]{figs-appendix-arxiv/gemma\_basic\_reflexion\_last\_reflexion\_max.png}%
    \\[1ex]
    \caption{Performance comparison of Gemma-7B in the Basic scenario across different agent types. Each plot shows relative performance differences (bars) and normalized average scores (vertical lines) across all 23 Atari environments. Top row: Comparison between (left) CoT vs. Reflexion\_last, (middle) CoT vs. Reflexion\_max, and (right) Naive vs. CoT agents. Bottom row: Comparison between (left) Naive vs. Reflexion\_last, (middle) Naive vs. Reflexion\_max, and (right) Reflexion\_last vs. Reflexion\_max agents.}
\end{figure}

\begin{figure}[htb!]
    \centering
    \includegraphics[width=0.32\textwidth]{figs-appendix-arxiv/gemma\_obscuring\_cot\_reflexion\_last.png}%
    \includegraphics[width=0.32\textwidth]{figs-appendix-arxiv/gemma\_obscuring\_cot\_reflexion\_max.png}%
    \includegraphics[width=0.32\textwidth]{figs-appendix-arxiv/gemma\_obscuring\_naive\_cot.png}%
    \\[1ex]
    \includegraphics[width=0.32\textwidth]{figs-appendix-arxiv/gemma\_obscuring\_naive\_reflexion\_last.png}%
    \includegraphics[width=0.32\textwidth]{figs-appendix-arxiv/gemma\_obscuring\_naive\_reflexion\_max.png}%
    \includegraphics[width=0.32\textwidth]{figs-appendix-arxiv/gemma\_obscuring\_reflexion\_last\_reflexion\_max.png}%
    \\[1ex]
    \caption{Performance comparison of Gemma-7B in the Obscured scenario across different agent types. Each plot shows relative performance differences (bars) and normalized average scores (vertical lines) across all 23 Atari environments. Top row: Comparison between (left) CoT vs. Reflexion\_last, (middle) CoT vs. Reflexion\_max, and (right) Naive vs. CoT agents. Bottom row: Comparison between (left) Naive vs. Reflexion\_last, (middle) Naive vs. Reflexion\_max, and (right) Reflexion\_last vs. Reflexion\_max agents.}
\end{figure}

\begin{figure}[htb!]
    \centering
    \includegraphics[width=0.32\textwidth]{figs-appendix-arxiv/gemma\_game\_manual\_cot\_reflexion\_last.png}%
    \includegraphics[width=0.32\textwidth]{figs-appendix-arxiv/gemma\_game\_manual\_cot\_reflexion\_max.png}%
    \includegraphics[width=0.32\textwidth]{figs-appendix-arxiv/gemma\_game\_manual\_naive\_cot.png}%
    \\[1ex]
    \includegraphics[width=0.32\textwidth]{figs-appendix-arxiv/gemma\_game\_manual\_naive\_reflexion\_last.png}%
    \includegraphics[width=0.32\textwidth]{figs-appendix-arxiv/gemma\_game\_manual\_naive\_reflexion\_max.png}%
    \includegraphics[width=0.32\textwidth]{figs-appendix-arxiv/gemma\_game\_manual\_reflexion\_last\_reflexion\_max.png}%
    \\[1ex]
    \caption{Performance comparison of Gemma-7B in the Game Manual scenario across different agent types. Each plot shows relative performance differences (bars) and normalized average scores (vertical lines) across all 23 Atari environments. Top row: Comparison between (left) CoT vs. Reflexion\_last, (middle) CoT vs. Reflexion\_max, and (right) Naive vs. CoT agents. Bottom row: Comparison between (left) Naive vs. Reflexion\_last, (middle) Naive vs. Reflexion\_max, and (right) Reflexion\_last vs. Reflexion\_max agents.}
\end{figure}

\begin{figure}[htb!]
    \centering
    \includegraphics[width=0.32\textwidth]{figs-appendix-arxiv/gemma\_RL\_traj\_cot\_reflexion\_last.png}%
    \includegraphics[width=0.32\textwidth]{figs-appendix-arxiv/gemma\_RL\_traj\_cot\_reflexion\_max.png}%
    \includegraphics[width=0.32\textwidth]{figs-appendix-arxiv/gemma\_RL\_traj\_naive\_cot.png}%
    \\[1ex]
    \includegraphics[width=0.32\textwidth]{figs-appendix-arxiv/gemma\_RL\_traj\_naive\_reflexion\_last.png}%
    \includegraphics[width=0.32\textwidth]{figs-appendix-arxiv/gemma\_RL\_traj\_naive\_reflexion\_max.png}%
    \includegraphics[width=0.32\textwidth]{figs-appendix-arxiv/gemma\_RL\_traj\_reflexion\_last\_reflexion\_max.png}%
    \\[1ex]
    \caption{Performance comparison of Gemma-7B in the RL Trajectory scenario across different agent types. Each plot shows relative performance differences (bars) and normalized average scores (vertical lines) across all 23 Atari environments. Top row: Comparison between (left) CoT vs. Reflexion\_last, (middle) CoT vs. Reflexion\_max, and (right) Naive vs. CoT agents. Bottom row: Comparison between (left) Naive vs. Reflexion\_last, (middle) Naive vs. Reflexion\_max, and (right) Reflexion\_last vs. Reflexion\_max agents.}
\end{figure}

\begin{figure}[htb!]
    \centering
    \includegraphics[width=0.32\textwidth]{figs-appendix-arxiv/basic\_naive\_llama\_gemma.png}%
    \includegraphics[width=0.32\textwidth]{figs-appendix-arxiv/basic\_naive\_qwen7b\_gemma.png}%
    \includegraphics[width=0.32\textwidth]{figs-appendix-arxiv/basic\_naive\_qwen7b\_llama.png}%
    \\[1ex]
   \includegraphics[width=0.32\textwidth]{figs-appendix-arxiv/basic\_cot\_llama\_gemma.png}%
    \includegraphics[width=0.32\textwidth]{figs-appendix-arxiv/basic\_cot\_qwen7b\_gemma.png}%
    \includegraphics[width=0.32\textwidth]{figs-appendix-arxiv/basic\_cot\_qwen7b\_llama.png}%
    \\[1ex]
    \caption{Cross-model performance comparison in the Basic scenario using Naive agent (top row) and Chain-of-Thought (CoT) agent (bottom row). Each plot shows relative performance differences (bars) and normalized average scores (vertical lines) across all 23 Atari environments. Top row: Comparison between (left) Llama3.1-8B vs. Gemma-7B, (middle) Qwen2.5-7B vs. Gemma-7B, and (right) Qwen2.5-7B vs. Llama3.1-8B using the Naive agent. Bottom row: Same model comparisons using the CoT agent.}
\end{figure}

\begin{figure}[htb!]
    \centering
    \includegraphics[width=0.32\textwidth]{figs-appendix-arxiv/basic\_reflexion\_last\_llama\_gemma.png}%
    \includegraphics[width=0.32\textwidth]{figs-appendix-arxiv/basic\_reflexion\_last\_qwen7b\_gemma.png}%
    \includegraphics[width=0.32\textwidth]{figs-appendix-arxiv/basic\_reflexion\_last\_qwen7b\_llama.png}%
    \\[1ex]
    \includegraphics[width=0.32\textwidth]{figs-appendix-arxiv/basic\_reflexion\_max\_llama\_gemma.png}%
    \includegraphics[width=0.32\textwidth]{figs-appendix-arxiv/basic\_reflexion\_max\_qwen7b\_gemma.png}%
    \includegraphics[width=0.32\textwidth]{figs-appendix-arxiv/basic\_reflexion\_max\_qwen7b\_llama.png}%
    \\[1ex]
    \caption{Cross-model performance comparison in the Basic scenario using Reflexion\_last agent (top row) and Reflexion\_max agent (bottom row). Each plot shows relative performance differences (bars) and normalized average scores (vertical lines) across all 23 Atari environments. Top row: Comparison between (left) Llama3.1-8B vs. Gemma-7B, (middle) Qwen2.5-7B vs. Gemma-7B, and (right) Qwen2.5-7B vs. Llama3.1-8B using the Reflexion\_last agent. Bottom row: Same model comparisons using the Reflexion\_max agent.}
\end{figure}

\begin{figure}[htb!]
    \centering
    \includegraphics[width=0.32\textwidth]{figs-appendix-arxiv/obscuring\_naive\_llama\_gemma.png}%
    \includegraphics[width=0.32\textwidth]{figs-appendix-arxiv/obscuring\_naive\_qwen7b\_gemma.png}%
    \includegraphics[width=0.32\textwidth]{figs-appendix-arxiv/obscuring\_naive\_qwen7b\_llama.png}%
    \\[1ex]
    \includegraphics[width=0.32\textwidth]{figs-appendix-arxiv/obscuring\_cot\_llama\_gemma.png}%
    \includegraphics[width=0.32\textwidth]{figs-appendix-arxiv/obscuring\_cot\_qwen7b\_gemma.png}%
    \includegraphics[width=0.32\textwidth]{figs-appendix-arxiv/obscuring\_cot\_qwen7b\_llama.png}%
    \\[1ex]
    \caption{Cross-model performance comparison in the Obscured scenario using Naive agent (top row) and Chain-of-Thought (CoT) agent (bottom row). Each plot shows relative performance differences (bars) and normalized average scores (vertical lines) across all 23 Atari environments. Top row: Comparison between (left) Llama3.1-8B vs. Gemma-7B, (middle) Qwen2.5-7B vs. Gemma-7B, and (right) Qwen2.5-7B vs. Llama3.1-8B using the Naive agent. Bottom row: Same model comparisons using the CoT agent.}
\end{figure}

\begin{figure}[htb!]
    \centering
    \includegraphics[width=0.32\textwidth]{figs-appendix-arxiv/obscuring\_reflexion\_last\_llama\_gemma.png}%
    \includegraphics[width=0.32\textwidth]{figs-appendix-arxiv/obscuring\_reflexion\_last\_qwen7b\_gemma.png}%
    \includegraphics[width=0.32\textwidth]{figs-appendix-arxiv/obscuring\_reflexion\_last\_qwen7b\_llama.png}%
    \\[1ex]
    \includegraphics[width=0.32\textwidth]{figs-appendix-arxiv/obscuring\_reflexion\_max\_llama\_gemma.png}%
    \includegraphics[width=0.32\textwidth]{figs-appendix-arxiv/obscuring\_reflexion\_max\_qwen7b\_gemma.png}%
    \includegraphics[width=0.32\textwidth]{figs-appendix-arxiv/obscuring\_reflexion\_max\_qwen7b\_llama.png}%
    \\[1ex]
    \caption{Cross-model performance comparison in the Obscured scenario using Reflexion\_last agent (top row) and Reflexion\_max agent (bottom row). Each plot shows relative performance differences (bars) and normalized average scores (vertical lines) across all 23 Atari environments. Top row: Comparison between (left) Llama3.1-8B vs. Gemma-7B, (middle) Qwen2.5-7B vs. Gemma-7B, and (right) Qwen2.5-7B vs. Llama3.1-8B using the Reflexion\_last agent. Bottom row: Same model comparisons using the Reflexion\_max agent.}
\end{figure}

\begin{figure}[htb!]
    \centering
    \includegraphics[width=0.32\textwidth]{figs-appendix-arxiv/game\_manual\_naive\_llama\_gemma.png}%
    \includegraphics[width=0.32\textwidth]{figs-appendix-arxiv/game\_manual\_naive\_qwen7b\_gemma.png}%
    \includegraphics[width=0.32\textwidth]{figs-appendix-arxiv/game\_manual\_naive\_qwen7b\_llama.png}%
    \\[1ex]
    \includegraphics[width=0.32\textwidth]{figs-appendix-arxiv/game\_manual\_cot\_llama\_gemma.png}%
    \includegraphics[width=0.32\textwidth]{figs-appendix-arxiv/game\_manual\_cot\_qwen7b\_gemma.png}%
    \includegraphics[width=0.32\textwidth]{figs-appendix-arxiv/game\_manual\_cot\_qwen7b\_llama.png}%
    \\[1ex]
    \caption{Cross-model performance comparison in the Game Manual scenario using Naive agent (top row) and Chain-of-Thought (CoT) agent (bottom row). Each plot shows relative performance differences (bars) and normalized average scores (vertical lines) across all 23 Atari environments. Top row: Comparison between (left) Llama3.1-8B vs. Gemma-7B, (middle) Qwen2.5-7B vs. Gemma-7B, and (right) Qwen2.5-7B vs. Llama3.1-8B using the Naive agent. Bottom row: Same model comparisons using the CoT agent.}
\end{figure}

\begin{figure}[htb!]
    \centering
    \includegraphics[width=0.32\textwidth]{figs-appendix-arxiv/game\_manual\_reflexion\_last\_llama\_gemma.png}%
    \includegraphics[width=0.32\textwidth]{figs-appendix-arxiv/game\_manual\_reflexion\_last\_qwen7b\_gemma.png}%
    \includegraphics[width=0.32\textwidth]{figs-appendix-arxiv/game\_manual\_reflexion\_last\_qwen7b\_llama.png}%
    \\[1ex]
    \includegraphics[width=0.32\textwidth]{figs-appendix-arxiv/game\_manual\_reflexion\_max\_llama\_gemma.png}%
    \includegraphics[width=0.32\textwidth]{figs-appendix-arxiv/game\_manual\_reflexion\_max\_qwen7b\_gemma.png}%
    \includegraphics[width=0.32\textwidth]{figs-appendix-arxiv/game\_manual\_reflexion\_max\_qwen7b\_llama.png}%
    \\[1ex]
    \caption{Cross-model performance comparison in the Game Manual scenario using Reflexion\_last agent (top row) and Reflexion\_max agent (bottom row). Each plot shows relative performance differences (bars) and normalized average scores (vertical lines) across all 23 Atari environments. Top row: Comparison between (left) Llama3.1-8B vs. Gemma-7B, (middle) Qwen2.5-7B vs. Gemma-7B, and (right) Qwen2.5-7B vs. Llama3.1-8B using the Reflexion\_last agent. Bottom row: Same model comparisons using the Reflexion\_max agent.}
\end{figure}

\begin{figure}[htb!]
    \centering
    \includegraphics[width=0.32\textwidth]{figs-appendix-arxiv/RL\_traj\_naive\_llama\_gemma.png}%
    \includegraphics[width=0.32\textwidth]{figs-appendix-arxiv/RL\_traj\_naive\_qwen7b\_gemma.png}%
    \includegraphics[width=0.32\textwidth]{figs-appendix-arxiv/RL\_traj\_naive\_qwen7b\_llama.png}%
    \\[1ex]
    \includegraphics[width=0.32\textwidth]{figs-appendix-arxiv/RL\_traj\_cot\_llama\_gemma.png}%
    \includegraphics[width=0.32\textwidth]{figs-appendix-arxiv/RL\_traj\_cot\_qwen7b\_gemma.png}%
    \includegraphics[width=0.32\textwidth]{figs-appendix-arxiv/RL\_traj\_cot\_qwen7b\_llama.png}%
    \\[1ex]
    \caption{Cross-model performance comparison in the RL Trajectory scenario using Naive agent (top row) and Chain-of-Thought (CoT) agent (bottom row). Each plot shows relative performance differences (bars) and normalized average scores (vertical lines) across all 23 Atari environments. Top row: Comparison between (left) Llama3.1-8B vs. Gemma-7B, (middle) Qwen2.5-7B vs. Gemma-7B, and (right) Qwen2.5-7B vs. Llama3.1-8B using the Naive agent. Bottom row: Same model comparisons using the CoT agent.}
\end{figure}

\begin{figure}[htb!]
    \centering
    \includegraphics[width=0.32\textwidth]{figs-appendix-arxiv/RL\_traj\_reflexion\_last\_llama\_gemma.png}%
    \includegraphics[width=0.32\textwidth]{figs-appendix-arxiv/RL\_traj\_reflexion\_last\_qwen7b\_gemma.png}%
    \includegraphics[width=0.32\textwidth]{figs-appendix-arxiv/RL\_traj\_reflexion\_last\_qwen7b\_llama.png}%
    \\[1ex]
    \includegraphics[width=0.32\textwidth]{figs-appendix-arxiv/RL\_traj\_reflexion\_max\_llama\_gemma.png}%
    \includegraphics[width=0.32\textwidth]{figs-appendix-arxiv/RL\_traj\_reflexion\_max\_qwen7b\_gemma.png}%
    \includegraphics[width=0.32\textwidth]{figs-appendix-arxiv/RL\_traj\_reflexion\_max\_qwen7b\_llama.png}%
    \\[1ex]
    \caption{Cross-model performance comparison in the RL Trajectory scenario using Reflexion\_last agent (top row) and Reflexion\_max agent (bottom row). Each plot shows relative performance differences (bars) and normalized average scores (vertical lines) across all 23 Atari environments. Top row: Comparison between (left) Llama3.1-8B vs. Gemma-7B, (middle) Qwen2.5-7B vs. Gemma-7B, and (right) Qwen2.5-7B vs. Llama3.1-8B using the Reflexion\_last agent. Bottom row: Same model comparisons using the Reflexion\_max agent.}
\end{figure}

\subsection{Key Observations}

The visualization structure allows us to highlight several important experimental design elements. 
The comparative analysis between Basic, Obscured, Game Manual, and Reference-based scenarios reveals how different forms of prior knowledge impact performance. 
By comparing these scenarios across models and agent types, we can isolate the relative importance of domain-specific vocabulary, explicit rule knowledge, and expert demonstrations in guiding language model decision-making.

The comparison between Naive, CoT, Reflexion\_last, and Reflexion\_max agents helps determine whether explicit reasoning strategies and reflection mechanisms provide consistent benefits across different games and knowledge conditions. 
This analysis is particularly important for understanding how different forms of prompted reasoning affect long-horizon decision-making capabilities.

By comparing Qwen2.5-7B, Llama3.1-8B, and Gemma-7B under identical conditions, we can examine whether architectural differences lead to systematic performance variations in long-horizon sequential decision-making. 
These comparisons allow us to assess whether certain model architectures are inherently better suited to particular types of reasoning required by different Atari games.

The visualization of performance across all 23 Atari environments highlights game-specific challenges and reveals which combinations of models, scenarios, and agent frameworks are most effective for particular types of games. 
This fine-grained analysis helps identify specific strengths and weaknesses across different experimental configurations.

\subsection{Methodological Considerations}

These extended results should be interpreted with several methodological considerations in mind. 
The relative performance differences across individual games demonstrate the inherent variability in sequential decision-making tasks, suggesting that no single approach is universally optimal across all game environments. 
This variability underscores the importance of evaluating language agent performance across diverse task types.

All comparative analyses use one configuration as a baseline, allowing for direct assessment of relative improvements or degradations. 
This approach enables meaningful comparisons across different experimental conditions while controlling for game-specific factors that might otherwise confound interpretation.

As noted in the main text, the substantial computational demands of these experiments  necessitated certain practical limitations, including a reduced evaluation horizon of 1,000 steps rather than the full 100,000 steps. 
While this reduction allows for broader experimental coverage, it may not fully capture the challenges of extremely long-horizon reasoning.

Performance on Atari games should be considered in the context of the specific challenges they present (spatial reasoning, planning, partial observability, and temporal reasoning) rather than as a general measure of language model capability. 
These games were selected specifically because they exercise different aspects of decision-making that are relevant to long-horizon planning.

The comprehensive nature of these visualizations allows for nuanced analysis of the interplay between model architectures, knowledge conditions, and prompting strategies in long-horizon sequential decision-making tasks. 
These visualizations provide valuable qualitative insights into the relative strengths and weaknesses of different approaches across the TextAtari benchmark. 
Together, they offer a foundation for understanding how different factors contribute to language agent performance on extended planning horizons.

\section{Border Impact}\label{sec:impact}

TextAtari introduces a benchmark for evaluating language agents on extremely long-horizon decision-making tasks, carrying various societal implications that warrant careful consideration. 
By establishing a rigorous evaluation standard for long-horizon reasoning, this benchmark may accelerate progress in temporal reasoning and strategic planning while highlighting the substantial gap between current AI systems and human capabilities. 
However, the framework's substantial computational demands raise important questions about research accessibility, environmental impact, and potential exacerbation of existing disparities in AI research.

Advances in long-horizon reasoning capabilities could transfer to beneficial applications across domains requiring extended planning, such as logistics optimization and healthcare coordination. 
Simultaneously, these same capabilities might enable more sophisticated autonomous systems for potentially harmful applications, including surveillance, automated disinformation campaigns, or autonomous weapons systems. 
There's also risk that optimizing for performance on game-based benchmarks could prioritize capabilities that don't transfer well to more nuanced real-world contexts involving ethical considerations, cultural sensitivity, or human welfare concerns.

TextAtari's findings regarding the performance gap between language agents and humans in extended reasoning tasks could inform more effective human-AI collaboration frameworks, potentially leading to more productive partnerships rather than full automation approaches in complex domains. 
The benchmark's design, which requires explicit reasoning traces in some agent configurations, encourages more interpretable AI systems and enables analysis of how language models construct and maintain internal representations over time.

While TextAtari itself represents a controlled research environment with minimal direct risk, the capabilities it aims to measure and advance have significant implications for AI development trajectories. 
The techniques developed to improve performance could be applied in both beneficial and harmful contexts—enhancing assistive technologies for individuals with cognitive impairments, but also potentially enabling more sophisticated autonomous systems for cyber attacks or manipulation. 
We encourage ongoing ethical reflection and governance discussions regarding long-horizon reasoning in autonomous systems as this research area progresses.

\clearpage
\newpage

\bibliographystyleapp{icml2025}
\bibliographyapp{appendix}

\end{document}